\documentclass[runningheads]{llncs}

\usepackage{graphicx}
\usepackage{multirow,makecell}
\usepackage{enumitem}

\usepackage{subcaption}
\usepackage{comment}
\usepackage{amsmath,amssymb}
\usepackage{color}
\usepackage{url}
\usepackage{hyperref}
\usepackage[capitalize,nameinlink]{cleveref}
\usepackage{graphicx}
\usepackage{booktabs}
\usepackage{pifont}
\usepackage{rotating}

\newif\ifreview
\reviewfalse

\ifreview
	\usepackage{lineno}

	\linenumbers
\fi

\begin{document}

\def\GCPRTrack{Fast Review Track}

\newcommand{\xmark}{\ding{55}}
\title{G3DST: Generalizing 3D Style Transfer with Neural Radiance Fields across Scenes and Styles}

% CAMERA READY SUBMISSION
\titlerunning{G3DST}

\author{Adil Meric \and
Umut Kocasari \and
Matthias Nie{\ss}ner \and
    Barbara Roessle
    }

\authorrunning{A. Meric et al.}

\institute{Technical University of Munich, Munich, Germany \\
\email{\{adil.meric, umut.kocasar, niessner, barbara.roessle\}@tum.de}}

\maketitle 
\vspace{-0.8cm}
\begin{figure}
    \centering % <-- added
\begin{subfigure}{0.27\textwidth}
  \includegraphics[width=\linewidth]{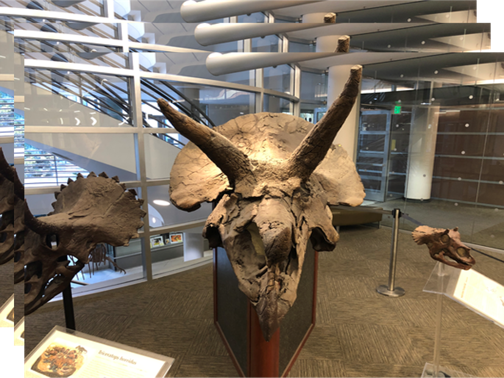}
\end{subfigure}
\begin{subfigure}{0.27\textwidth}
  \includegraphics[width=\linewidth]{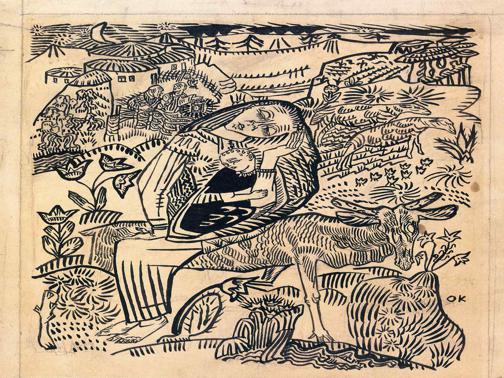}
\end{subfigure} % <-- added
\begin{subfigure}{0.27\textwidth}
  \includegraphics[width=\linewidth]{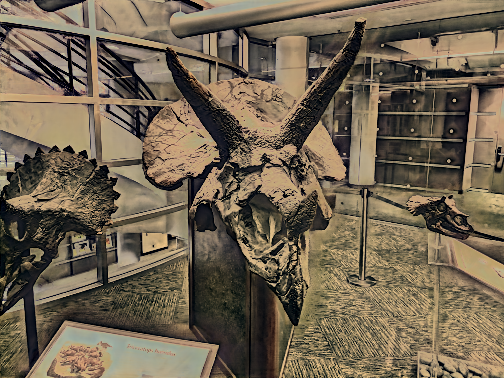}
\end{subfigure} % <-- added
\medskip

\begin{subfigure}{0.27\textwidth}
  \includegraphics[width=\linewidth]{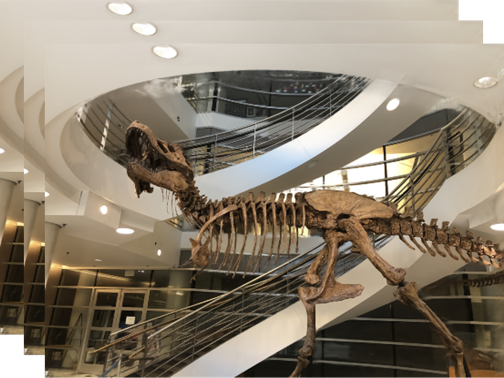}
  \caption{Source Views}
\end{subfigure}
\begin{subfigure}{0.27\textwidth}
  \includegraphics[width=\linewidth]{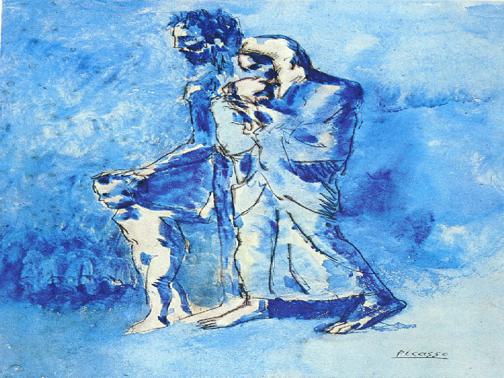}
  \caption{Style Image}
\end{subfigure}
\begin{subfigure}{0.27\textwidth}
  \includegraphics[width=\linewidth]{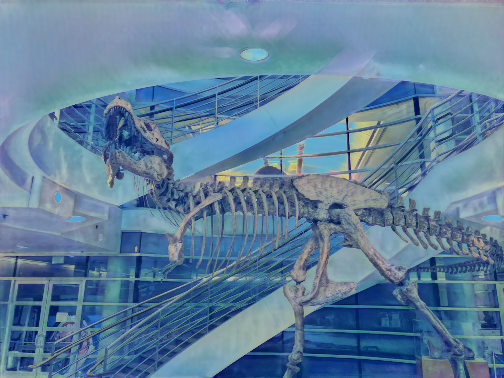}
  \caption{Stylized Novel View}
\end{subfigure}
\smallskip

\vspace{-0.2cm}
\caption{\textbf{Generalizable 3D Style Transfer.} Given a set of source views and a style image, our method renders view-consistent, stylized novel views without any per-scenew or per-style optimization.
}
\vspace{-0.8cm}
\label{fig:teaser}
\end{figure}

\begin{abstract}
%\vspace{-0.3cm}
Neural Radiance Fields (NeRF) have emerged as a powerful tool for creating highly detailed and photorealistic scenes. Existing methods for NeRF-based 3D style transfer need extensive per-scene optimization for single or multiple styles, limiting the applicability and efficiency of 3D style transfer. In this work, we overcome the limitations of existing methods by rendering stylized novel views from a NeRF without the need for per-scene or per-style optimization. To this end, we take advantage of a generalizable NeRF model to facilitate style transfer in 3D, thereby enabling the use of a single learned model across various scenes. By incorporating a hypernetwork into a generalizable NeRF, our approach enables on-the-fly generation of stylized novel views. Moreover, we introduce a novel flow-based multi-view consistency loss to preserve consistency across multiple views. We evaluate our method across various scenes and artistic styles and show its performance in generating high-quality and multi-view consistent stylized images without the need for a scene-specific implicit model. Our findings demonstrate that this approach not only achieves a good visual quality comparable to that of per-scene methods but also significantly enhances efficiency and applicability, marking a notable advancement in the field of 3D style transfer.

%\vspace{-0.1cm}
\end{abstract}

\clearpage
%
%
%
%%%%%%%%% BODY TEXT
\section{Introduction}

\label{sec:intro}

Style transfer is a method that transfers the style of a target style image onto a content image while preserving the structure of the content image. Traditional methods \cite{gatys2016image} solve this problem by iteratively optimizing style and content losses together. Iterative methods require optimization from scratch for each source and style image. On top of that, optimization-based methods take significant amount of time because gradients are calculated for each pixel at every step. Stylization networks speed up this process by applying style transfer in a single forward pass. Existing methods \cite{huang2017arbitrary,NIPS2017_49182f81,li2018learning,Svoboda_2020_CVPR} demonstrate valuable performance on the 2D domain without requiring per-target optimization.

Recent methods apply style transfer techniques to 3D domain. Given a set of source views and a target style image, 3D style transfer aims to generate high-quality stylized novel views for a 3D scene while preserving the consistency across novel views. Pixels corresponding to the same point in 3D space across different views should have similar colors without considering the lighting effect. Therefore, applying style transfer to a single frame should utilize information from all the other frames or the 3D structure represented by all frames. Simply applying 2D style transfer methods to rendered novel-views produces highly stylized images. However, as noted in a recent study \cite{chiang2022stylizing}, the results are highly inconsistent. Existing NeRF-based 3D style transfer methods \cite{chiang2022stylizing,chen2022upst,zhang2022arf,Liu_2023_CVPR}, produce consistent high-quality stylization. 

Hyper~\cite{chiang2022stylizing}, applies style transfer in two steps: geometric training and stylization training. In the geometric training phase, the geometry branch of the framework is trained by using the input images similar to NeRF training. Then, the geometry layers are freezed and stylization layers of the framework are trained by utilizing a style image dataset. Afterwards, novel views with unseen style images can be produced for the trained scene. To apply the style transfer to a new scene, all training phases need to be redone. StyleRF \cite{Liu_2023_CVPR}, utilizes a similar two-stage training schema. StyleRF employs TensoRF \cite{tensorf} as an implicit geometry network, the geometric training phase of StyleRF is much faster compared to Hyper; however, it also requires re-training for every new scene. This method uses content transformation for the features of the points on the ray and applies style transformation to the feature map produced by volume rendering.

All of these methods require per-scene optimization and some of them require per-style optimization which is time-consuming and costly. Some other techniques~\cite{yu2021pixelnerf,wang2021ibrnet,t2023is} make NeRF generalizable and produce novel views without per-scene overfitting. GNT~\cite{t2023is} utilizes transformers to learn the feature aggregation from different views, and the rendering operation by itself: a view transformer aggregates features from neighbouring source views to calculate relevant features for the target view; a ray transformer performs the rendering by fusing features along camera rays to produce pixel colors. Both are trained to minimize the photometric loss between the produced image and the ground truth image. After training, GNT synthesizes novel views from unseen scenes on the fly. Hence, generalizable NeRF models produce novel views from unseen images but they cannot apply 3D style transfer. To the best of our knowledge, there is no existing method for generalizable style transfer on 3D scenes.

In this work, we propose a network for 3D style transfer that can generalize to novel scenes and styles on the fly. We take advantage of a generalizable NeRF approach and develop a stylization network whose weights are predicted by a hypernetwork which is trained on diverse style images and flexibly reacts to novel styles. 
Our feed-forward framework enables both novel-view synthesis and 3D style transfer at inference time in a very time efficient manner. We further introduce a multi-view consistency loss which effectively preserves stylization quality and consistency across novel views (\cref{fig:teaser,fig:result}). We demonstrate that our generalizable method achieves stylization results which outperform recent approaches~\cite{chiang2022stylizing,Liu_2023_CVPR} that require time-consuming per-scene optimization.

The contributions of this work are summarized as follows:
\begin{enumerate}
    \item We introduce the first 3D style transfer method that generalizes across scenes and styles.
    \item We propose a consistency loss using optical flow to preserve multi-view consistency during generalizable, NeRF-based 3D style transfer.
\end{enumerate}

\section{Related Work}
\label{sec:rw}

\begin{figure}[t]
  \centering
  \includegraphics[width=1\textwidth]{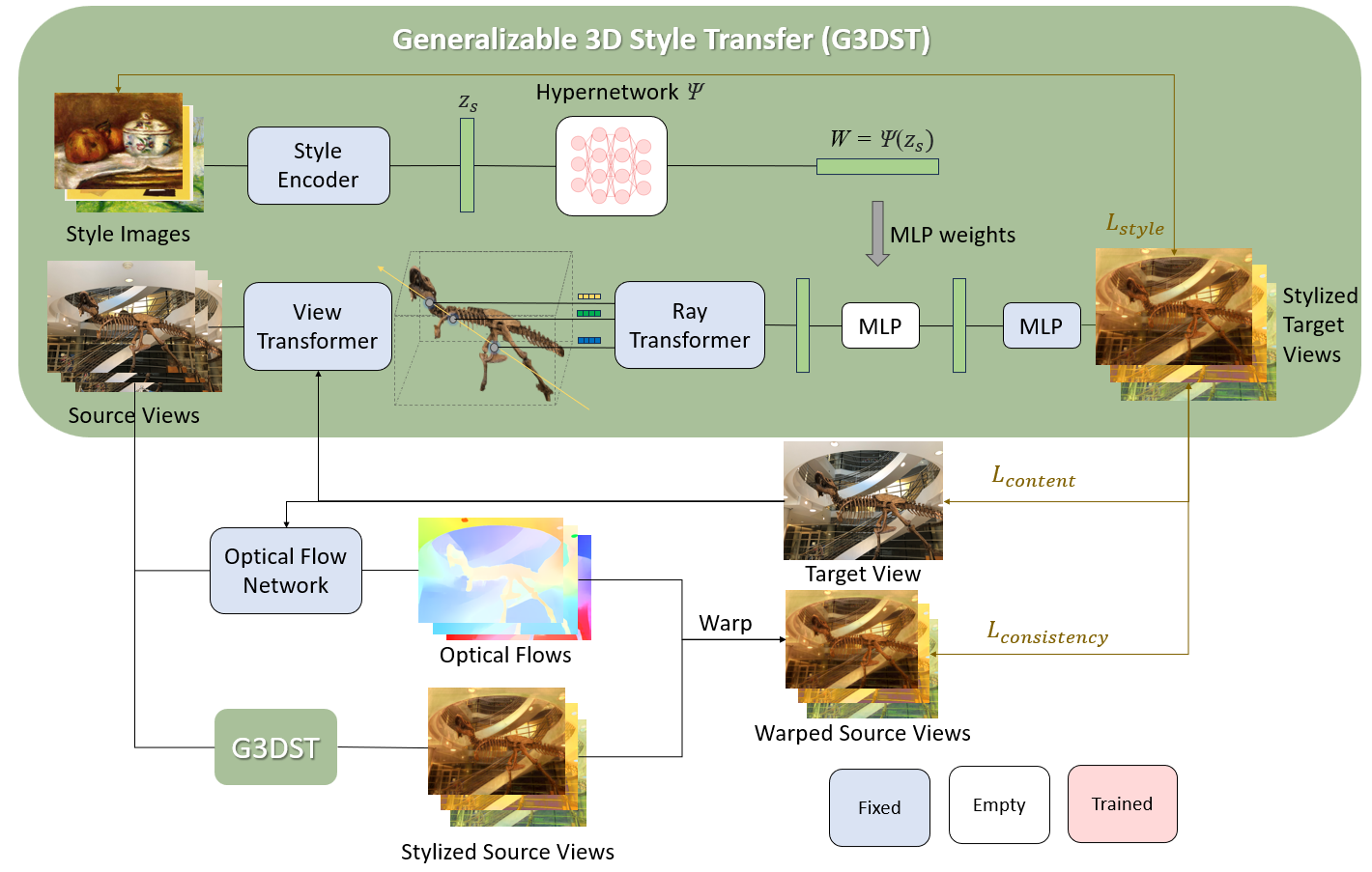}
  \caption{\textbf{Framework.} We utilize a hypernetwork to apply a style transformation to the features of a generalizable transformer-based NeRF. The hypernetwork takes a style latent vector $z_s$ as input and outputs weights and biases of an intermediate MLP, which stylizes the aggregated ray features. This operation is repeated for each ray in the image to produce a high quality stylized image. We calculate the optical flow between source views and minimize the difference between corresponding pixels in stylized images.
  }
  \label{fig:framework}
  \vspace{-0.5cm}
\end{figure}

\subsubsection{Image Style Transfer.} Gatys et al.\ \cite{gatys2016image} first introduced neural style transfer on the image domain. Image style transfer can be grouped into feed-forward-based~\cite{huang2017arbitrary,johnson2016perceptual,ulyanov2016texture} and optimization-based~\cite{gatys2016image,chiu2020iterative,gatys2017controlling} approaches. These methods take advantage of pretrained convolutional neural networks, such as VGG~\cite{simonyan2014very}, to extract image features for the content and the style loss. Content loss is commonly used to maintain the features of the input image. At the same time, a style loss often computes a Gram matrix to encourage similar feature statistics to the style image~\cite{johnson2016perceptual}. 
While neural style transfer on images achieves impressive stylization results, the application in the video or 3D domain remains challenging. 

\subsubsection{Video Style Transfer.} Video style transfer methods address the task of applying a consistent style to a sequence of RGB frames. These works can similarly be clustered into feed-forward-based~\cite{chen2017coherent,chen2020optical,gao2018reconet,gao2020fast,gupta2017characterizing,ReReVST2020,xia2021real} and optimization-based~\cite{ruder2016artistic,ruder2018artistic} approaches. Directly applying 2D style transfer on each video frame typically leads to temporal inconsistency and flickering. To prevent such artifacts, video style transfer employs temporal coherency or optical flow constraints. Inspired by this, our method introduces optical flow constraints to feed-forward-based 3D style transfer, however, instead of relying on image style transfer to stylize individual views and aggregating them, we take a multi-view approach already in the stylization, leading to a fairly consistent result to start with. 

\subsubsection{Neural Scene Representations.}
3D scene representations based on implicit functions~\cite{chen2019learning,mescheder2019occupancy,niemeyer2020differentiable,yariv2020multiview} employ neural networks to store the properties of a scene. These representations support differentiable rendering, hence can be optimized from multi-view images. In particular, Neural Radiance Fields (NeRF)~\cite{mildenhall2020nerf} have shown impressive, highly detailed novel view synthesis results on complex 3D scenes. NeRF represents a scene as an implicit function, realized as an MLP, which maps a 3D coordinate and viewing direction to a color and density. Volume rendering aggregates colors and densities along target camera rays to obtain pixel colors in the target view. 
% per-scene approaches
% hybrid and grid NeRFs
Follow-up works have extended NeRF to voxel grids~\cite{fridovich2022plenoxels,liu2020neural,sun2022direct}, decomposed tensors~\cite{chan2022efficient,kplanes_2023,tensorf} or hash maps~\cite{mueller2022instant} to increase optimization and rendering speed. 
Nonetheless, these works do not generalize across scenes and per-scene optimization is still needed. 
% generalizable NeRFs
Another line of works achieve generalizable NeRFs~\cite{yu2021pixelnerf,wang2021ibrnet,t2023is,SRF,mvsnerf} by using a feature-based, generic view interpolation function. Given multiple source views as input, they synthesize novel views in a feed-forward manner without per-scene fitting.
% GNT
Generalizable NeRF Transformer (GNT) \cite{t2023is} employs transformers to render novel views from input images: a view transformer aggregates multi-view features that project onto a target camera ray; these features are combined along target camera rays using a ray transformer and the resulting features are mapped to RGB with an MLP. 
% meaning for 3D style transfer
Through their high-fidelity reconstructions, NeRFs are very interesting representations for 3D style transfer. Aiming for feed-forward 3D style transfer that is generalizable across scenes, our method builds up on GNT and extends its transformer architecture with a hypernetwork to further achieve generalization to novel styles. 

\subsubsection{3D Style Transfer.}
3D style transfer stylizes an entire 3D scene according to a style image, such that renderings from novel viewpoints consistently follow the style of the style image. 
Prior works apply 3D style transfer on point clouds~\cite{huang2021learning,mu20223d} or meshes~\cite{yin20213dstylenet,Michel_2022_CVPR,hollein2022stylemesh} as a 3D scene representation. The quality of these approaches is often bounded by a limited geometric accuracy or the requirement for a given reconstruction. 
To circumvent these shortcomings, recent approaches take advantage of NeRF as a high quality scene representation for 3D style transfer~\cite{chen2022upst,chiang2022stylizing,nguyenphuoc2022snerf,zhang2022arf,fan2022unified,Huang22StylizedNeRF}. SNeRF~\cite{nguyenphuoc2022snerf} and ARF~\cite{zhang2022arf} optimize a NeRF to render novel views that match the style of a style image. While these methods achieve high-quality stylization, they involve time-intensive per-scene and per-style optimization. 
% optimized for multiple styles
Other NeRF-based approaches~\cite{fan2022unified,Huang22StylizedNeRF} are optimized for a collection of style images using latent embeddings to identify them, however, they still cannot generalize to novel style images. 
In contrast, \cite{chen2022upst,chiang2022stylizing} achieve generalization across styles by employing a hypernetwork, that modifies the appearance layers of a NeRF representation based on encoded style images. 
Likewise, the recent work StyleRF~\cite{Liu_2023_CVPR} generalizes across styles by applying a style-dependent, adaptive transformation to the feature grid extracted from NeRF. Concurrent to our work, ConRF~\cite{miao2024conrf} and MM-NeRF~\cite{mmnerf} explore the usage of CLIP features for 3D style transfer that generalizes to novel styles in the form of images or text. While several methods are able to operate on unseen style images, none of them is able to generalize to a novel scene without retraining. We take it one step further and present a feed-forward-based 3D style transfer method that renders stylized views of novel scenes on the fly, without any per-scene or per-style optimization. 

\label{method}

\begin{figure}[!ht]
    \centering % <-- added
\begin{subfigure}{0.18\textwidth}
  \includegraphics[width=\linewidth]{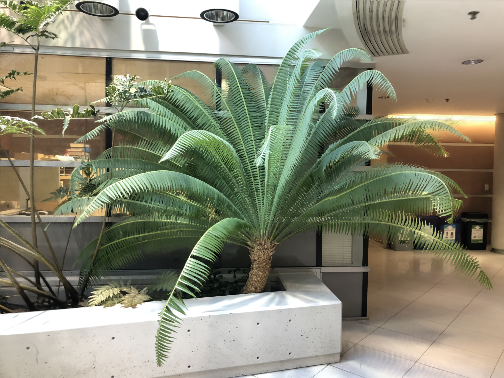}
\end{subfigure}
\begin{subfigure}{0.18\textwidth}
  \includegraphics[width=\linewidth]{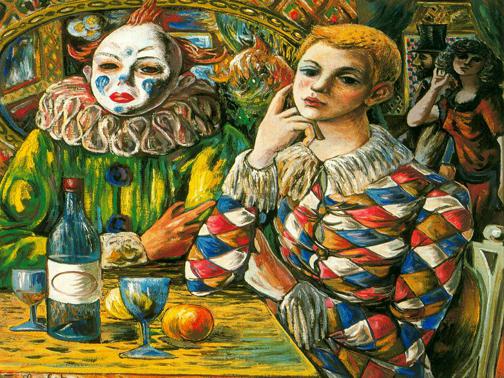}
\end{subfigure} % <-- added
\begin{subfigure}{0.18\textwidth}
  \includegraphics[width=\linewidth]{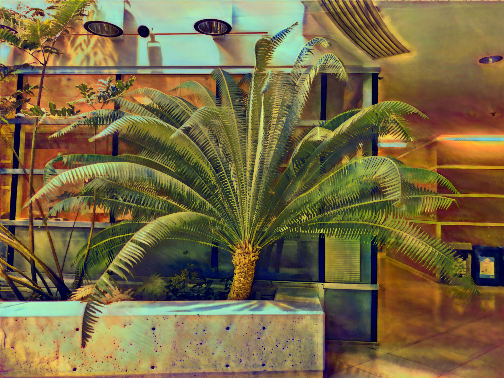}
\end{subfigure}
\begin{subfigure}{0.18\textwidth}
  \includegraphics[width=\linewidth]{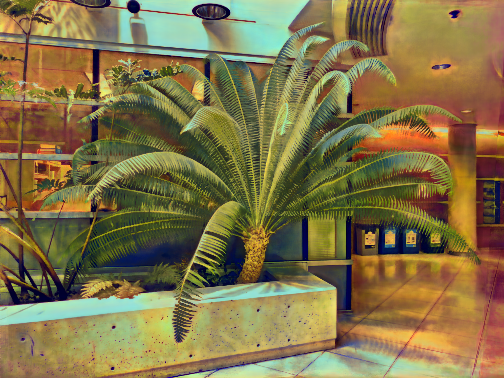}
\end{subfigure} % <-- added
\begin{subfigure}{0.18\textwidth}
  \includegraphics[width=\linewidth]{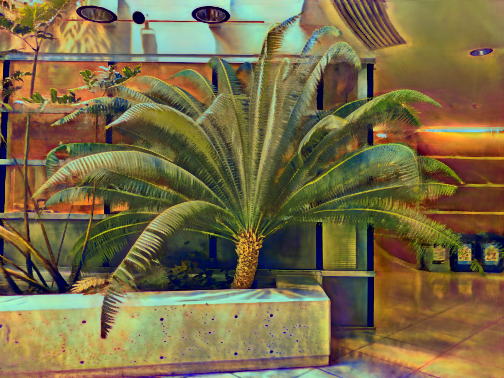}
\end{subfigure} % <-- added
\medskip

\begin{subfigure}{0.18\textwidth}
  \includegraphics[width=\linewidth]{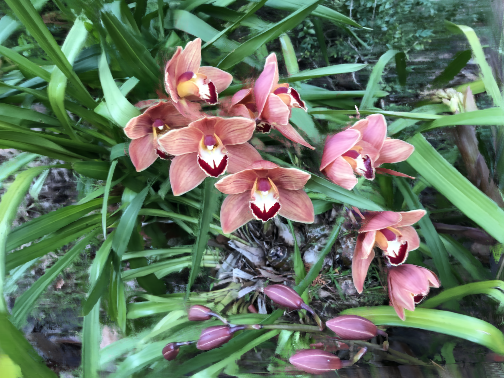}
\end{subfigure}
\begin{subfigure}{0.18\textwidth}
  \includegraphics[width=\linewidth]{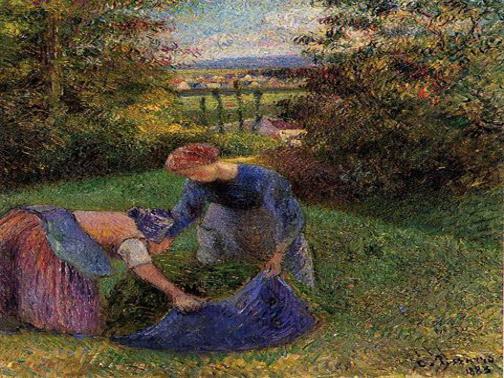}
\end{subfigure} % <-- added
\begin{subfigure}{0.18\textwidth}
  \includegraphics[width=\linewidth]{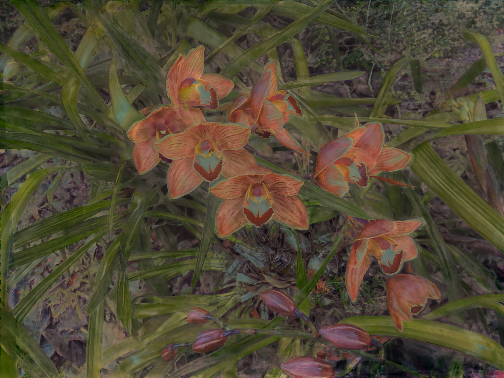}
\end{subfigure}
\begin{subfigure}{0.18\textwidth}
  \includegraphics[width=\linewidth]{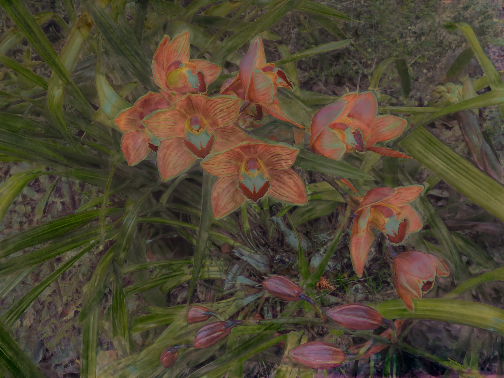}
\end{subfigure} % <-- added
\begin{subfigure}{0.18\textwidth}
  \includegraphics[width=\linewidth]{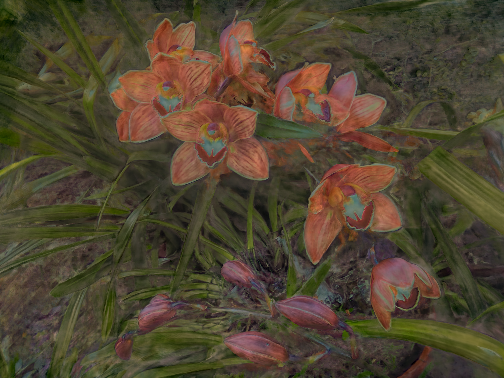}
\end{subfigure} % <-- added
\medskip

\begin{subfigure}{0.18\textwidth}
  \includegraphics[width=\linewidth]{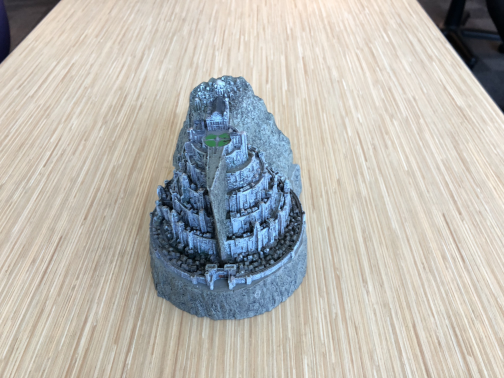}
\end{subfigure}
\begin{subfigure}{0.18\textwidth}
  \includegraphics[width=\linewidth]{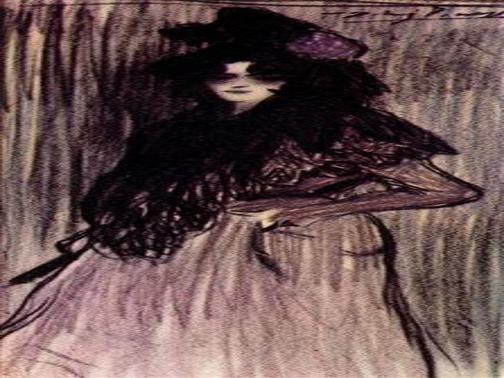}
\end{subfigure} % <-- added
\begin{subfigure}{0.18\textwidth}
  \includegraphics[width=\linewidth]{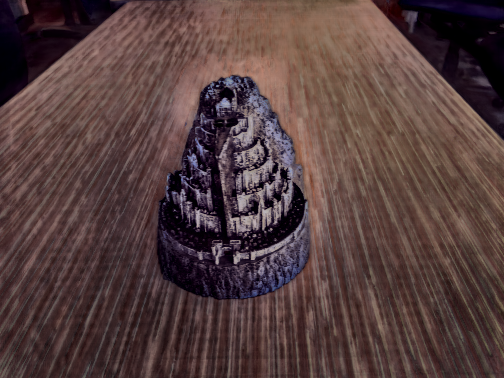}
\end{subfigure}
\begin{subfigure}{0.18\textwidth}
  \includegraphics[width=\linewidth]{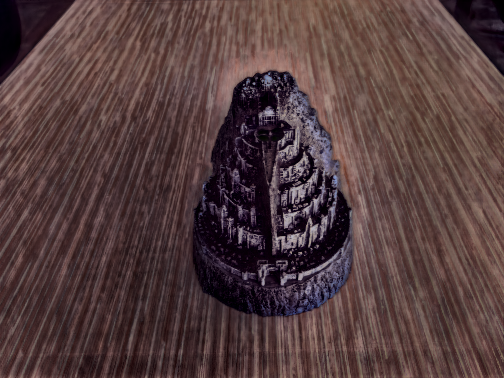}
\end{subfigure} % <-- added
\begin{subfigure}{0.18\textwidth}
  \includegraphics[width=\linewidth]{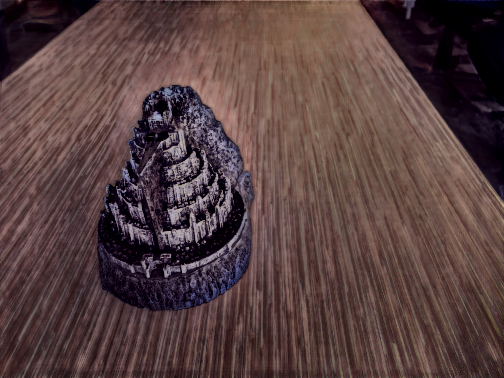}
\end{subfigure} % <-- added
\medskip

\begin{subfigure}{0.18\textwidth}
  \includegraphics[width=\linewidth]{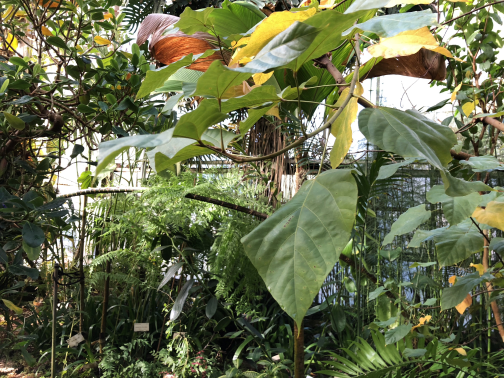}
\end{subfigure}
\begin{subfigure}{0.18\textwidth}
  \includegraphics[width=\linewidth]{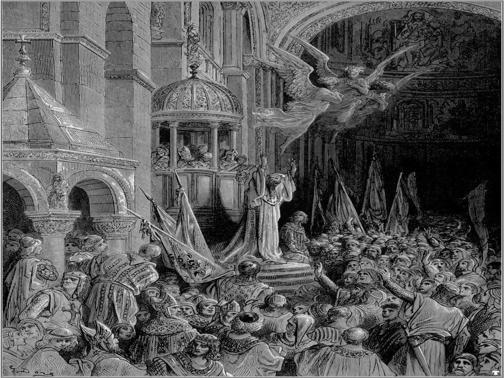}
\end{subfigure}
\begin{subfigure}{0.18\textwidth}
  \includegraphics[width=\linewidth]{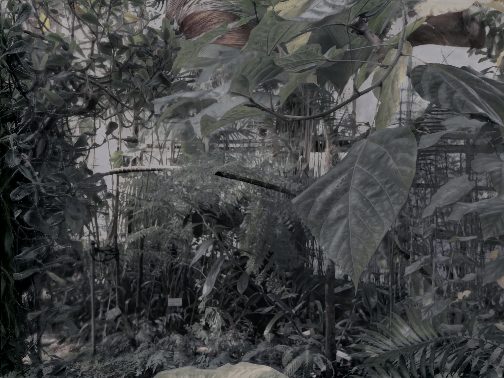}
\end{subfigure}
\begin{subfigure}{0.18\textwidth}
  \includegraphics[width=\linewidth]{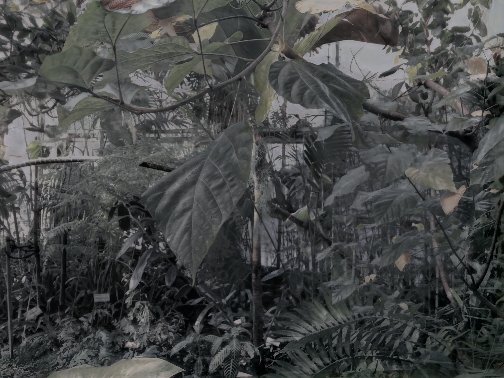}
\end{subfigure}
\begin{subfigure}{0.18\textwidth}
  \includegraphics[width=\linewidth]{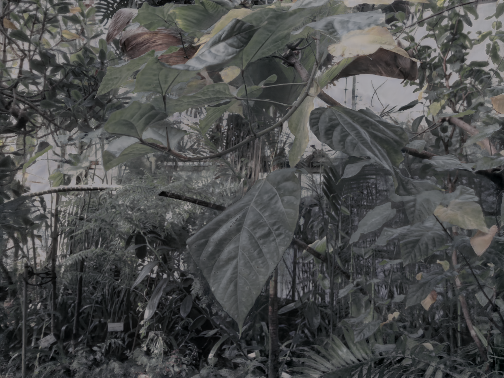}
\end{subfigure}
\medskip

\begin{subfigure}{0.18\textwidth}
  \includegraphics[width=\linewidth]{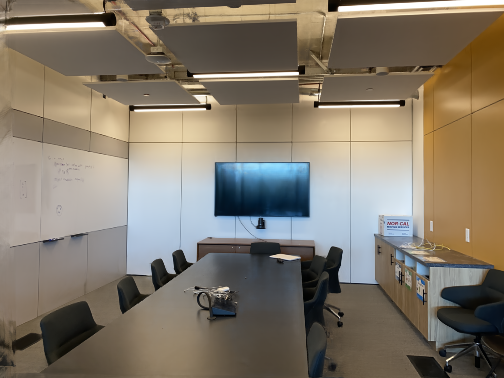}
  \caption{Scene}
\end{subfigure}
\begin{subfigure}{0.18\textwidth}
  \includegraphics[width=\linewidth]{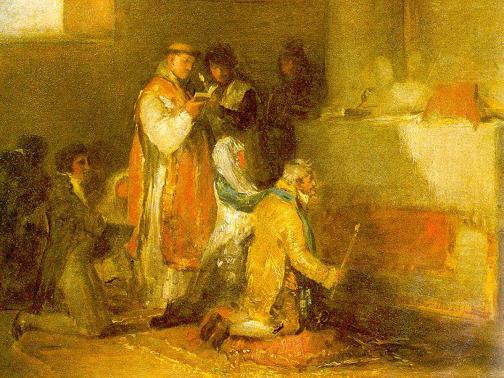}
  \caption{Style}
\end{subfigure}
\begin{subfigure}{0.18\textwidth}
  \includegraphics[width=\linewidth]{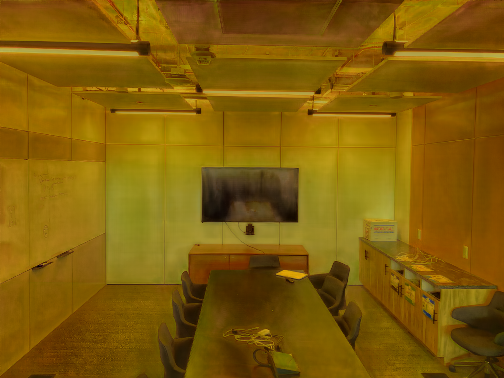}
  \caption{View 1}
\end{subfigure}
\begin{subfigure}{0.18\textwidth}
  \includegraphics[width=\linewidth]{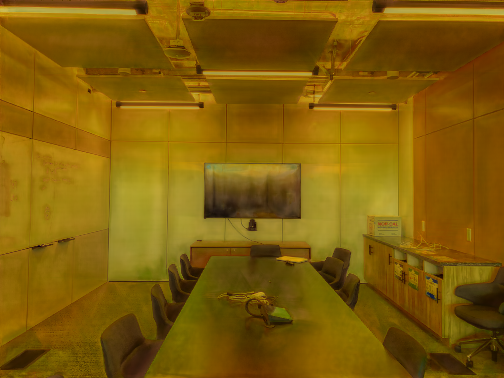}
  \caption{View 2}
\end{subfigure}
\begin{subfigure}{0.18\textwidth}
  \includegraphics[width=\linewidth]{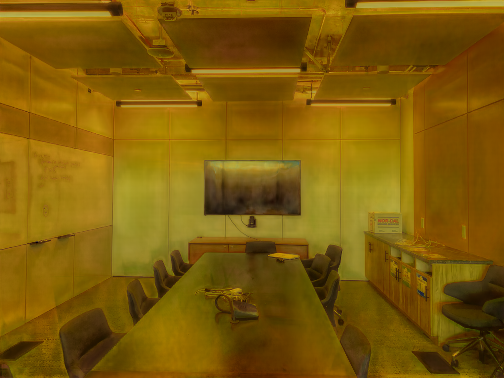}
  \caption{View 3}
\end{subfigure}
\medskip

\caption{\textbf{Results.} Our method captures the style and preserves view-consistency. The output images produced from different viewpoints are geometrically consistent and capture the stylistic details of the given style image.}
\label{fig:result}
\end{figure}
\begin{figure}[!ht]
    \centering % <-- added
\begin{subfigure}{0.18\textwidth}
  \includegraphics[width=\linewidth]{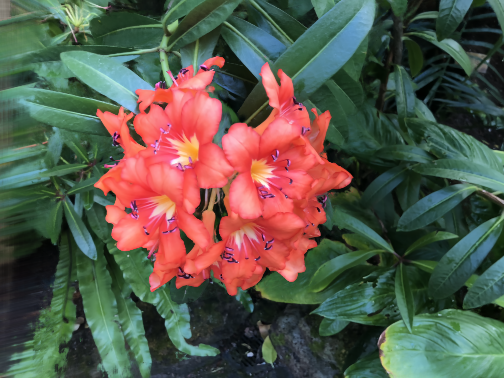}
\end{subfigure}
\begin{subfigure}{0.18\textwidth}
  \includegraphics[width=\linewidth]{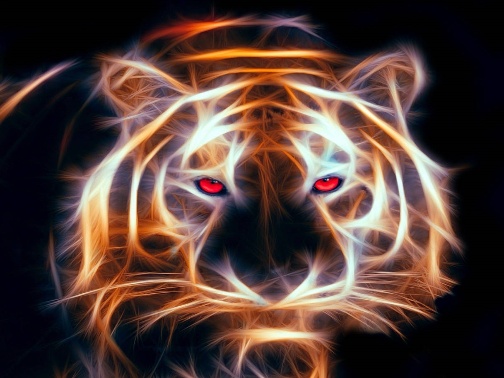}
\end{subfigure}
\begin{subfigure}{0.18\textwidth}
  \includegraphics[width=\linewidth]{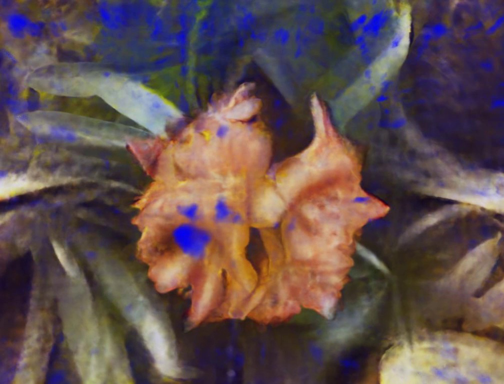}
\end{subfigure} % <-- added
\begin{subfigure}{0.18\textwidth}
  \includegraphics[width=\linewidth]{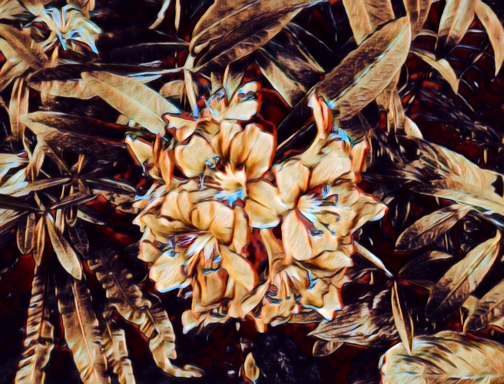}
\end{subfigure} % <-- added
\begin{subfigure}{0.18\textwidth}
  \includegraphics[width=\linewidth]{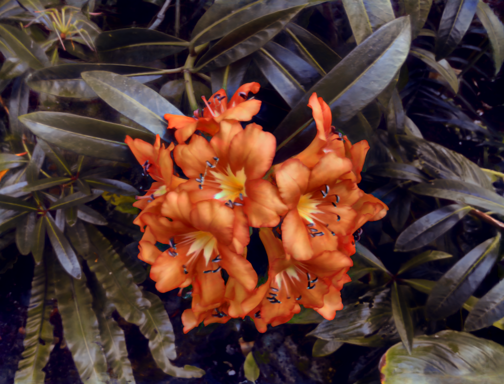}
\end{subfigure} % <-- added
\medskip

\iffalse %Close second flower figure
\begin{subfigure}{0.18\textwidth}
  \includegraphics[width=\linewidth]{./figures/results/flower.png}
\end{subfigure}
\begin{subfigure}{0.18\textwidth}
  \includegraphics[width=\linewidth]{figures/comparison/Port-en-Bessin,Sunday.jpg}
\end{subfigure}
\begin{subfigure}{0.18\textwidth}
  \includegraphics[width=\linewidth]{./figures/comparison/hyper_flower2.png}
\end{subfigure} % <-- added
\begin{subfigure}{0.18\textwidth}
  \includegraphics[width=\linewidth]{./figures/comparison/stylerf_flower2.png}
\end{subfigure} % <-- added
\begin{subfigure}{0.18\textwidth}
  \includegraphics[width=\linewidth]{./figures/comparison/Port-en-Bessin,Sunday_flower.png}
\end{subfigure} % <-- added
\medskip
\fi

\begin{subfigure}{0.18\textwidth}
  \includegraphics[width=\linewidth]{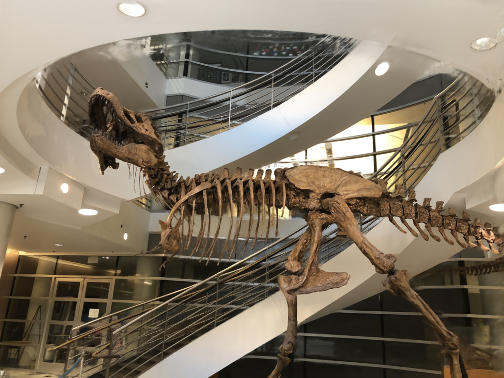}
\end{subfigure}
\begin{subfigure}{0.18\textwidth}
  \includegraphics[width=\linewidth]{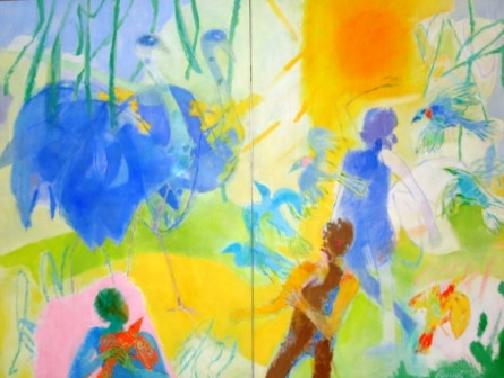}
\end{subfigure}
\begin{subfigure}{0.18\textwidth}
  \includegraphics[width=\linewidth]{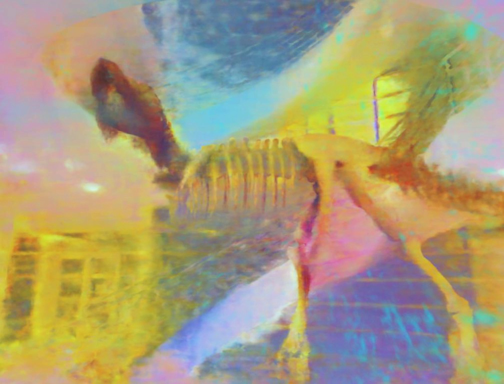}
\end{subfigure} % <-- added
\begin{subfigure}{0.18\textwidth}
  \includegraphics[width=\linewidth]{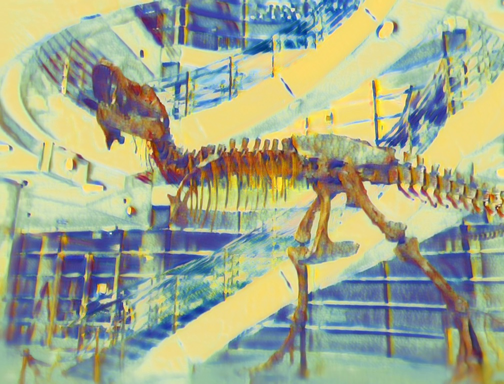}
\end{subfigure} % <-- added
\begin{subfigure}{0.18\textwidth}
  \includegraphics[width=\linewidth]{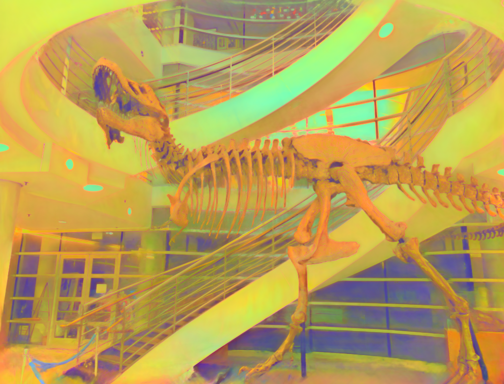}
\end{subfigure} % <-- added
\medskip

\begin{subfigure}{0.18\textwidth}
  \includegraphics[width=\linewidth]{./figures/results/trex.png}
\end{subfigure}
\begin{subfigure}{0.18\textwidth}
  \includegraphics[width=\linewidth]{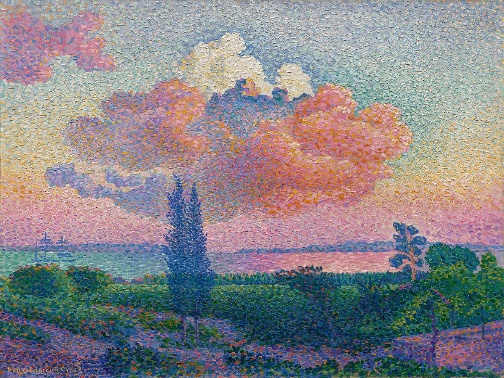}
\end{subfigure}
\begin{subfigure}{0.18\textwidth}
  \includegraphics[width=\linewidth]{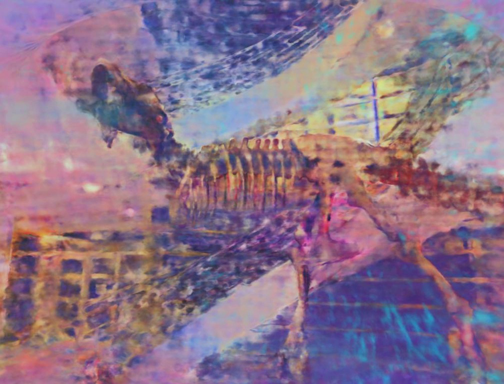}
\end{subfigure} % <-- added
\begin{subfigure}{0.18\textwidth}
  \includegraphics[width=\linewidth]{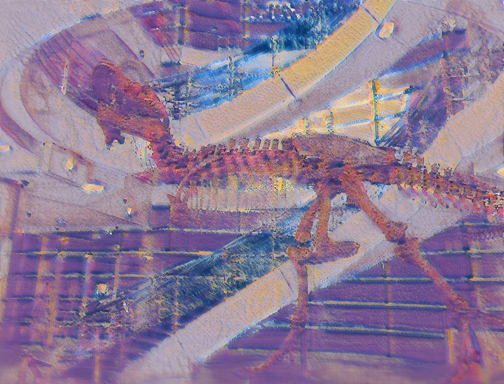}
\end{subfigure} % <-- added
\begin{subfigure}{0.18\textwidth}
  \includegraphics[width=\linewidth]{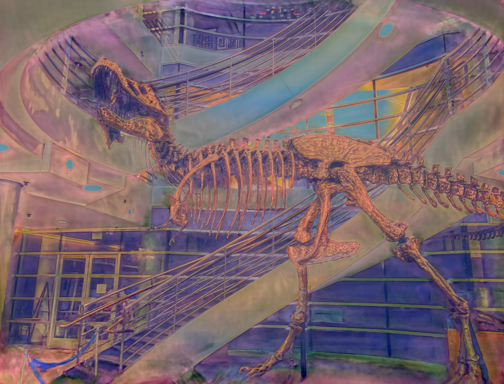}
\end{subfigure} % <-- added
\medskip

\begin{subfigure}{0.18\textwidth}
  \includegraphics[width=\linewidth]{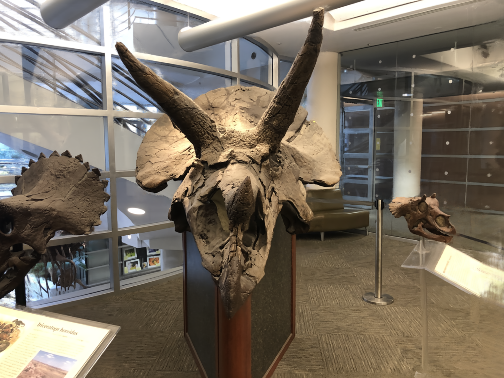}
\end{subfigure}
\begin{subfigure}{0.18\textwidth}
  \includegraphics[width=\linewidth]{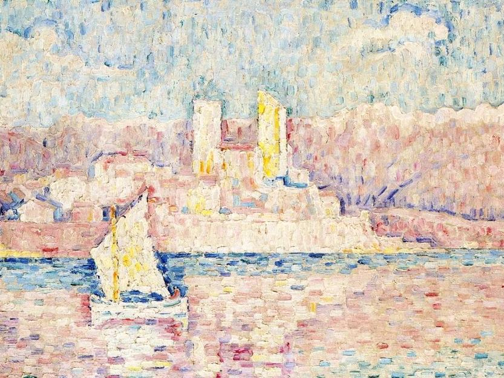}
\end{subfigure}
\begin{subfigure}{0.18\textwidth}
  \includegraphics[width=\linewidth]{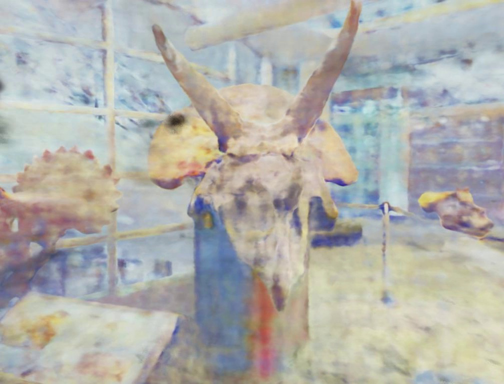}
\end{subfigure}
\begin{subfigure}{0.18\textwidth}
  \includegraphics[width=\linewidth]{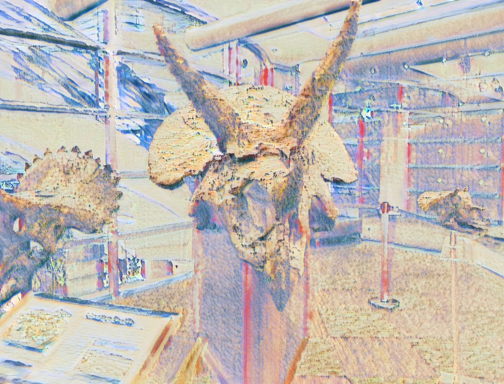}
\end{subfigure}
\begin{subfigure}{0.18\textwidth}
  \includegraphics[width=\linewidth]{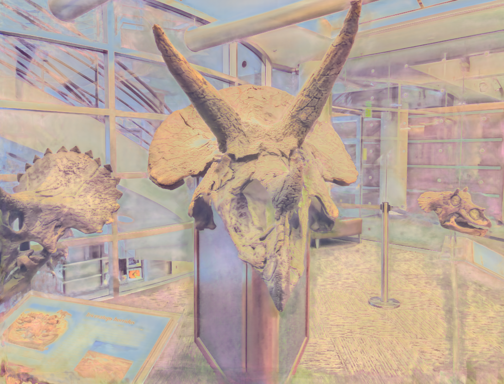}
\end{subfigure}
\medskip

\begin{subfigure}{0.18\textwidth}
  \includegraphics[width=\linewidth]{./figures/comparison/horns.png}
  \caption{Scene}
\end{subfigure}
\begin{subfigure}{0.18\textwidth}
  \includegraphics[width=\linewidth]{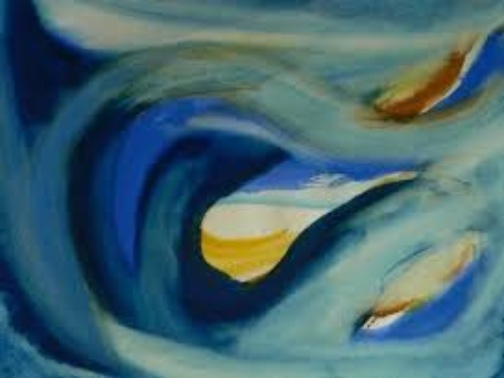}
  \caption{Style}
\end{subfigure}
\begin{subfigure}{0.18\textwidth}
  \includegraphics[width=\linewidth]{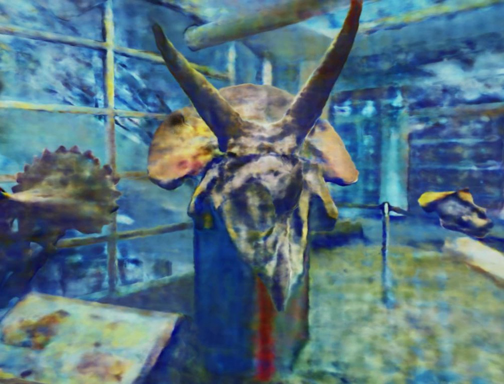}
  \caption{Hyper\cite{chiang2022stylizing}}
\end{subfigure}
\begin{subfigure}{0.18\textwidth}
  \includegraphics[width=\linewidth]{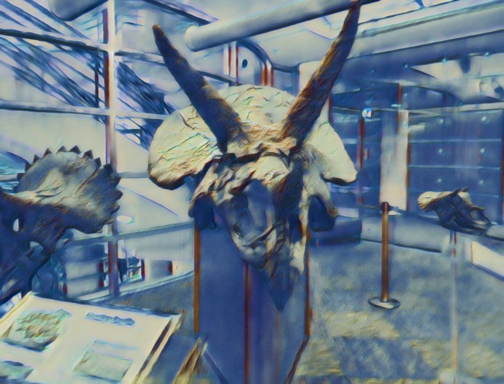}
  \caption{StyleRF\cite{Liu_2023_CVPR}}
\end{subfigure}
\begin{subfigure}{0.18\textwidth}
  \includegraphics[width=\linewidth]{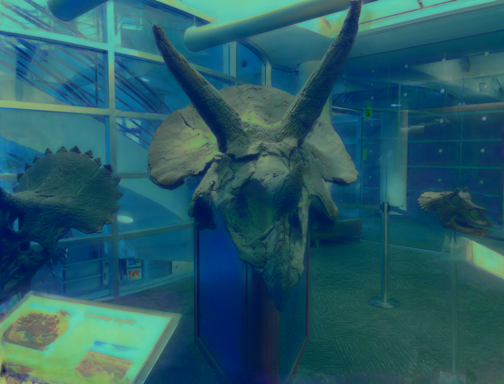}
  \caption{Ours}
\end{subfigure}
\smallskip

\caption{\textbf{Visual Comparison with Other Methods.} Hyper \cite{chiang2022stylizing} produces blurry results with artifacts, StyleRF \cite{Liu_2023_CVPR} captures the style and preserves the geometry in a more consistent way. Our method successfully captures the style of a given style image while preserving the geometric details.
}
\label{fig:comparison}
\end{figure}

\section{Method}
Our method enables the generation of stylistic novel views during inference by taking a single style image and a collection of scene images with their associated camera parameters as input. We propose a unified pipeline for 3D novel view synthesis and stylization during inference. Building on this, we introduce a novel approach for 3D style transfer by integrating a hypernetwork into a generalized 3D baseline. Furthermore, we enforce 3D consistency between different views with our novel multi-view consistency loss.
\subsection{Network Architecture} We adopt Generalizable NeRF Transformer (GNT) model \cite{t2023is} as our baseline for an implicit scene representation. As shown in Fig. \ref{fig:framework}, the geometric stage is structured around three primary components: view transformer, ray transformer, and a coloring MLP. The view transformer aggregates multi-view image features by incorporating epipolar geometry as an inductive bias to extract coordinate-aligned features from neighboring source views. More detailed, our geometric stage projects a 3D point into the image plane of input views and aggregates corresponding features using the view transformer for each sampled point on a ray. On top of that, the view-transformer utilizes epipolar geometry in its attention mechanism. Unlike volumetric rendering based methods, we employ a ray transformer for rendering. The ray transformer processes the extracted coordinate-aligned features for each ray. The output of the ray-transformer is passed to the coloring MLP which takes the processed ray as input and outputs a color $c \in \mathbb{R}^3$. Hence, no rendering equation is used. Our framework can be seen in Fig. \ref{fig:framework}.

In this work, we apply an intermediate feature transformation to the output of the ray transformer using an additional MLP as shown in Fig. \ref{fig:framework}. We develop a hypernetwork that estimates the weights of the intermediate MLP by conditioning on a latent vector $z_s$ of a style image. To produce style latents, we use a pretrained style-encoder of a Style VAE network, similar to \cite{chiang2022stylizing}. As given in Fig. \ref{fig:framework}, we only backpropagate through the hypernetwork during training, while keeping the geometry network and style-encoder frozen. By utilizing a hypernetwork on top of a generalizable baseline, we apply 3D style transfer to arbitrary scenes at inference time by only providing source images of a scene and a style image. In the consistency loss calculation, we estimate the optical flow between each view using the state-of-the-art optical flow estimation network RAFT~\cite{teed2020raft}. At training time, we stylize a selected source view and warp that view to the target view with the pre-calculated optical flow. We enforce consistency in the pixel space between the stylized target view and the stylized warped source view.

\subsection{Loss Functions}
To achieve stylization, while maintaining the content of the scene, we apply a content loss (Eq. \ref{eq:content}) and a style loss (Eq. \ref{eq:style}). On top of that, we employ a novel consistency loss (Eq. \ref{eq:consistency}) to preserve multi-view consistency across style-transfered images.
\subsubsection{Content \& Style Losses.}
We aim to change the style of a scene by preserving the geometric details of the scene. To preserve the geometric details, we use a content loss that minimizes the feature distance between a target image from a specific camera position, $I$, and its corresponding stylized image $\hat{I_s}$, the output of our method. 
\begin{equation}
\label{eq:content}
%\small
    L_{content} = \sum_j||\Phi_{j}(I) - \Phi_j(\hat{I_s}) ||^2_2
\end{equation}

where $\Phi$ is an ImageNet-pretrained VGG-19 network. 
$\Phi_j$ denotes the output obtained from the $j$th layer of the network.

We achieve style transfer by enforcing similarities between the feature statistics of the style image $I_s$, and the predicted stylized image $\hat{I_s}$ from our network. We use feature representations obtained by the first layer of the same VGG-19 network and measure the mean squared error (MSE) between feature means and standard deviations. 

\begin{equation}
\label{eq:style}
%\small
    \mathcal{L}_{style} = ||\mu(\Phi_{1}(I_s)) - \mu(\Phi_1(\hat{I_s})) ||^2_2 + ||\sigma(\Phi_{1}(I_s)) - \sigma(\Phi_1(\hat{I_s})) ||^2_2
\end{equation}
where $\mu$ and $\sigma$ denotes mean and standard deviation calculation. 

\subsubsection{Multi-View Consistency Loss.}
We define a novel multi-view consistency loss in the pixel domain. From the raw source images, we compute optical flow between all source image pairs using RAFT~\cite{teed2020raft}. During training we apply style transfer for two images, $\hat{I_s}^{(i)}$ and $\hat{I_s}^{(j)}$, warp the second image $\hat{I_s}^{(j)}$ with respect to the computed optical flow $F^{(j,i)}$, and calculate the masked MSE loss as stated in Eq. \ref{eq:consistency}. 

\begin{equation}
\label{eq:consistency}
%\small
    \mathcal{L}_{consistency} =  ||(M_{j,i} \odot \hat{I_s}^{(i)}) - (M_{j,i} \odot W(\hat{I_s}^{(j)}, F^{(j,i)})) ||^2_2
\end{equation}

where $F^{(j,i)}$ is the optical flow from view $j$ to view $i$, $W$ is the warping function, and $M_{j,i}$ is the pixel visibility mask after warping operation.
\\

The overall loss is defined as follows:
\begin{equation}
\label{eq:total}
%\small
    L_{total} = L_{content} + w_s L_{style} + w_c L_{consistency}
\end{equation}

where $w_c$ and $w_s$ balance the loss terms.

\begin{table}[t!]
\begin{center}
\resizebox{0.50\columnwidth}{!}{
\begin{tabular}{|c|c|c|}
\hline
\multirow{2}{*}{Method} & Generalization & Generalization \\
 & across Scenes & across Styles \\
\hline
Hyper \cite{chiang2022stylizing} & \xmark & \checkmark \\
StyleRF \cite{Liu_2023_CVPR} & \xmark & \checkmark\\
Ours & \checkmark & \checkmark  \\
\hline
\end{tabular}
}
\end{center}
\caption{\textbf{Generalization comparison.} Comparison of our method with state of the art methods. Recent methods such as Hyper and StyleRF generalize across style images; however, we are the only method that generalizes across scenes and styles, thereby eliminating the need for scene-specific training.}
\label{table:1}
\end{table}

\begin{table}[t!]
\begin{center}
\resizebox{0.9\columnwidth}{!}{
\begin{tabular}{|c|c|c|c|c|c|}
\hline
\multirow{2}{*}{Method} & Scene Independent & \multicolumn{2}{c|}{Scene Dependent Training} & Inference \\ \cline{3-4}
& Training & Geometry Training & Style Training & (per frame) \\
\hline
Hyper \cite{chiang2022stylizing} & - & 4 days & 2 days & 50s \\
StyleRF \cite{Liu_2023_CVPR} & - & 2 h 28 mins & 3h 10 mins & 18s\\
Ours & 1.5 days & - & - & 21s \\
\hline
\end{tabular}
} 
\end{center}
\caption{\textbf{Time comparison.} Comparison of our method to the methods that generalize across styles. Our method only requires 1.5 days of pre-training once on top of a generalizable NeRF network, and can produce new stylization results in seconds while other methods require hours of training for a new scene.}
\vspace{-0.7cm}
\label{table:2}
\end{table}

\section{Experiments}
\label{sec:exp}

\subsubsection{Dataset.} Our generalizable NeRF baseline \cite{t2023is} is trained with several datasets such as Google Scanned Object \cite{downs2022google}, RealEstate10K \cite{zhou2018stereo}, Spaces dataset \cite{flynn2019deepview} and real scenes from handheld cellphone captures \cite{mildenhall2019local,wang2021ibrnet}. In our experimental setup, we use the training set of captured real scenes from the handheld \cite{mildenhall2019local} and show our results on the validation set of Local Light Field Fusion (LLFF) dataset \cite{mildenhall2019local} which is not seen during the training of both our geometry baseline and stylization network. The training set of the stylization network consists of 42 scenes, totaling 1025 images, while the validation set involves 8 scenes with 355 images. For style images, we use the WikiArt dataset \cite{saleh2015large} and provided train/validation split which involves 57025 images for training, and 24421 images for validation. 
\subsubsection{Training Details.} At each training iteration, we select a random style image from our training set to increase the generalizability. We downsample our training data by a factor of 8 and conduct our experiments on 378x504 resolution which means at each iteration we shoot 378x504 rays. We sample 192 coarse points on a ray. We only apply style transformation to final ray feature, hence, this trained pipeline can be used with any desired resolution and number of samples along a ray. 'e initialize the hypernetwork so that it behaves like an identity function, which means regardless of the input it produces identity MLP weights. We trained our framework for 500 epochs using the Adam optimizer with learning rate $10^{-3}$ and a batch size of 16. 

\subsection{Results} Fig. \ref{fig:result} shows that the stylization results capture the style of the style image successfully. In the third row, we see that the stylized images are in claret red color and we can find the drawing style of the style image in the orchids results. Also, in the last row, the produced images have the characteristics of the style image in a clear way. At the same time, the image content is preserved and the color is consistent across different viewpoints.

\subsection{Baseline Comparisons}
We conduct detailed quantitative and qualitative analyses and compare our method with other style transfer methods, Hyper~\cite{chiang2022stylizing} and StyleRF~\cite{Liu_2023_CVPR}, which both generalize across styles but not across scenes. 

\begin{table}[t]
\begin{center}
\resizebox{0.7\columnwidth}{!}{
\begin{tabular}{|c|c|c|c|c|}
\hline
\multirow{2}{*}{Method} & \multicolumn{2}{c|}{Short Range Consistency} & \multicolumn{2}{c|}{Long Range Consistency} \\ \cline{2-5}
& LPIPS $\downarrow$ & RMSE $\downarrow$ & LPIPS $\downarrow$ & RMSE $\downarrow$ \\
\hline
Hyper \cite{chiang2022stylizing} & 0.1854 & 0.1136 & 0.2750 & 0.1635\\
StyleRF \cite{Liu_2023_CVPR} & 0.0637 & 0.0639 & 0.1624 & 0.1607\\
Ours & \textbf{0.0416} & \textbf{0.0435} & \textbf{0.1439} & \textbf{0.1168} \\
\hline
\end{tabular} 
}

\end{center}
\caption{\textbf{Consistency score comparison.} Comparison of our method with Hyper and StyleRF. Our method outperforms existing methods both in short range and long range consistency.}
\vspace{-0.5cm}
\label{table:3}
\end{table}

\begin{table}[t!]
\begin{center}
\resizebox{0.6\columnwidth}{!}{
\begin{tabular}{|c|c|c|}
\hline
Method & Stylization (1-5) $\uparrow$ & Consistency (1-5) $\uparrow$  \\
\hline
Hyper \cite{chiang2022stylizing}  & 1.86 & 1.50 \\
StyleRF \cite{Liu_2023_CVPR}  & \textbf{3.33} & \textbf{3.85}\\
Ours & 3.32 & 3.53 \\
\hline
\end{tabular}
}
\end{center}

\caption{\textbf{User study comparison.} Comparison of our method with recent works. Results are very close to StyleRF while outperforming Hyper. Considering StyleRF has an advantage due to per-scene overfitting, we reach similar scores without scene-dependent training.}
\vspace{-0.5cm}
\label{table:4}
\end{table}

\subsubsection{Generalization Comparison.}
\cref{table:1} compares the generalization capability of our method and the baselines. 
Given a set of source views from an arbitrary scene and an arbitrary style image, our method achieves high quality and consistent 3D style transfer in a generalizable manner. None of the existing methods generalize across both styles and scenes. SOTA methods such as Hyper \cite{chiang2022stylizing} and StyleRF \cite{Liu_2023_CVPR} generalize across styles; however, they require per-scene fitting. Our method calculates the feature transformation at inference time without requiring any extra optimization to the given scene or style image. This way, our framework stylizes unseen scenes with unseen styles almost instantly.

\begin{figure*}[t]
    \centering % <-- added 
    
\begin{subfigure}{0.19\textwidth}
  \includegraphics[width=\linewidth]{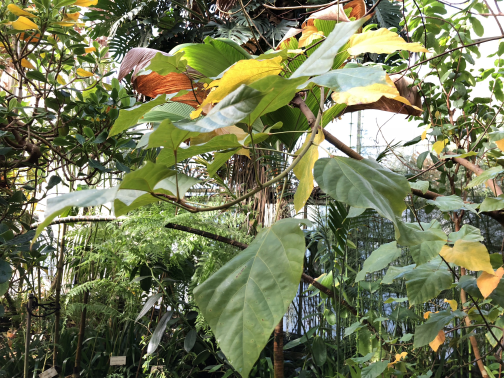}
\end{subfigure}
\begin{turn}{90}StyleRF\cite{Liu_2023_CVPR} \end{turn}
\begin{subfigure}{0.19\textwidth}
  \includegraphics[width=\linewidth]{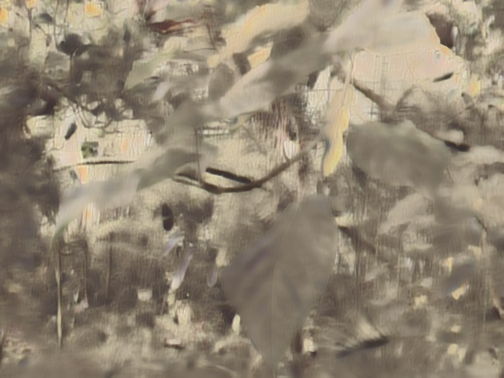}
\end{subfigure} % <-- added
\begin{subfigure}{0.19\textwidth}
  \includegraphics[width=\linewidth]{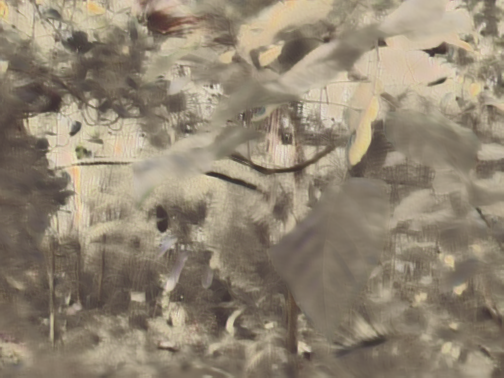}
\end{subfigure}
\begin{subfigure}{0.19\textwidth}
  \includegraphics[width=\linewidth]{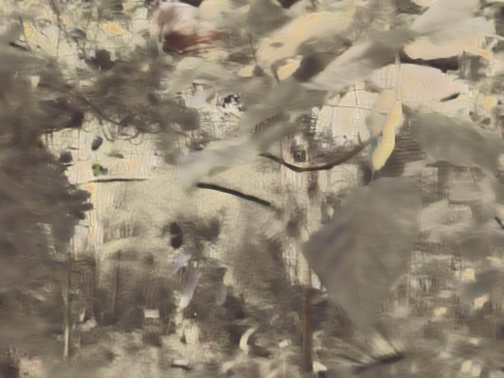}
\end{subfigure} % <-- added
\medskip

\begin{subfigure}{0.19\textwidth}
\includegraphics[width=\linewidth]{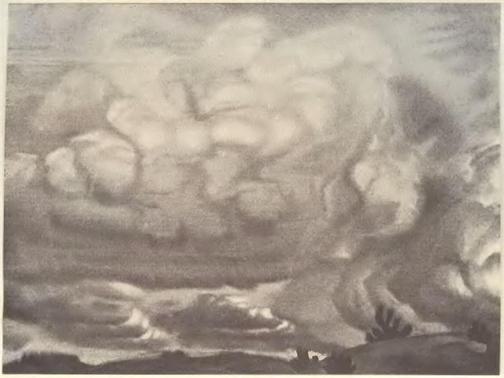}
\end{subfigure}
\begin{turn}{90} \; \; Ours \end{turn}
\begin{subfigure}{0.19\textwidth}
  \includegraphics[width=\linewidth]{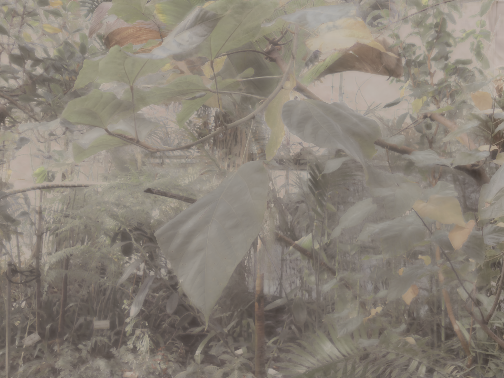}
\end{subfigure} % <-- added
\begin{subfigure}{0.19\textwidth}
  \includegraphics[width=\linewidth]{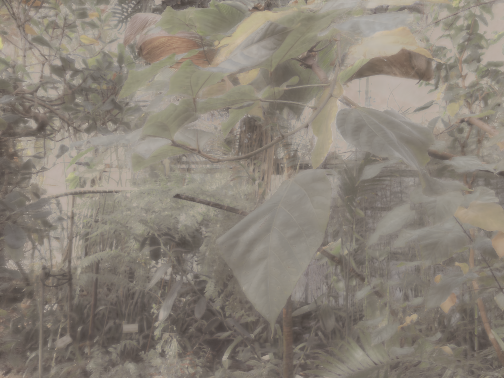}
\end{subfigure}
\begin{subfigure}{0.19\textwidth}
  \includegraphics[width=\linewidth]{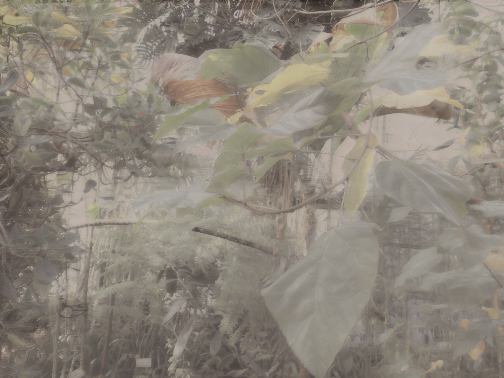}
\end{subfigure} % <-- added
\smallskip

 \textbf{--- --- ---  --- --- --- --- --- --- --- ---  --- --- --- --- --- --- --- --- --- --- ---}

\begin{subfigure}{0.19\textwidth}
  \includegraphics[width=\linewidth]
  {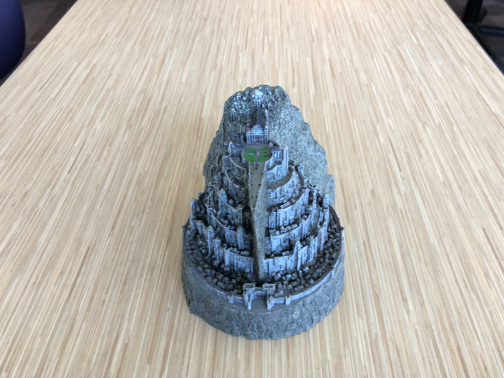}
\end{subfigure}
\begin{turn}{90}StyleRF\cite{Liu_2023_CVPR} \end{turn}
\begin{subfigure}{0.19\textwidth}
  \includegraphics[width=\linewidth]{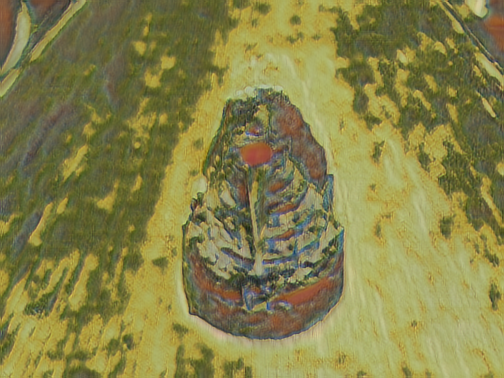}
\end{subfigure}
\begin{subfigure}{0.19\textwidth}
  \includegraphics[width=\linewidth]{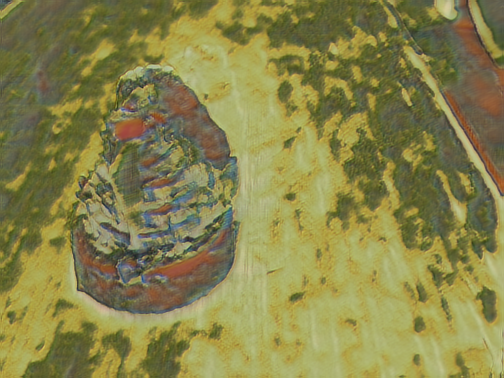}
\end{subfigure}
\begin{subfigure}{0.19\textwidth}
  \includegraphics[width=\linewidth]{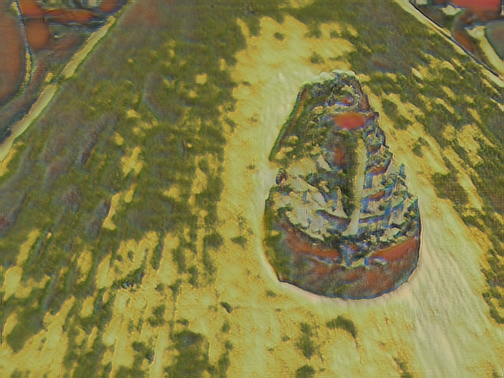}
\end{subfigure} % <-- added
\medskip

\begin{subfigure}{0.19\textwidth}
  \includegraphics[width=\linewidth]{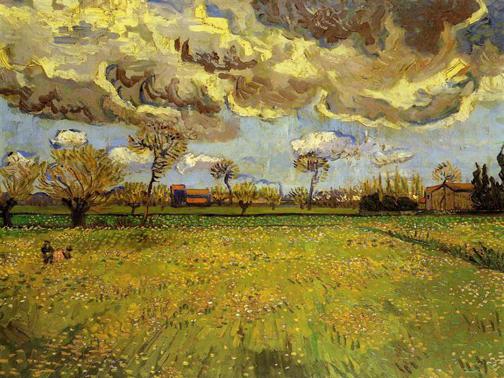}
  \caption{Scene/Style}
\end{subfigure}
\begin{turn}{90} \quad \; \; Ours \end{turn}
\begin{subfigure}{0.19\textwidth}
  \includegraphics[width=\linewidth]{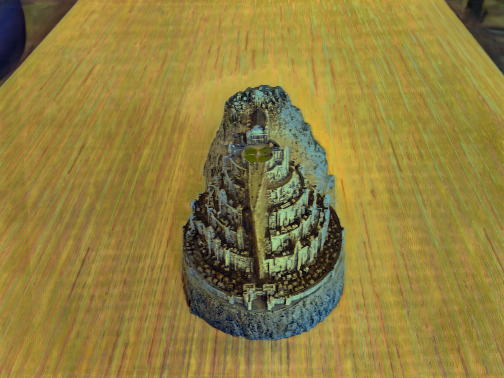}
  \caption{View 1}
\end{subfigure}
\begin{subfigure}{0.19\textwidth}
  \includegraphics[width=\linewidth]{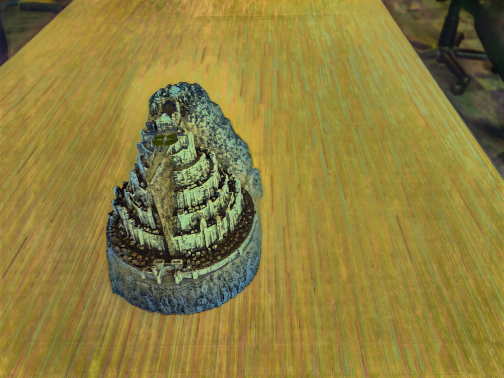}
  \caption{View 2}
\end{subfigure}
\begin{subfigure}{0.19\textwidth}
  \includegraphics[width=\linewidth]{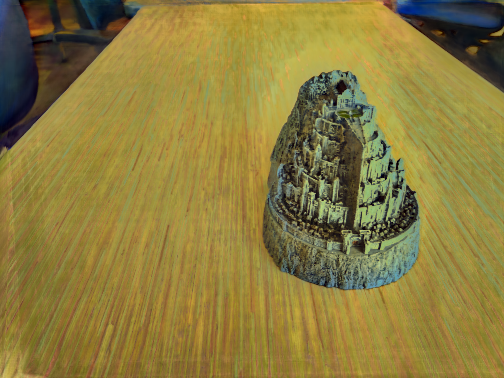}
  \caption{View 3}
\end{subfigure}
\smallskip

\caption{\textbf{Comparison of our method with StyleRF from different views.} Our method captures the style and preserves the geometry in different novel views. StyleRF cannot capture local geometric details, therefore, it struggles to preserve multi-view consistency.}
\label{fig:comparison_stylerf}
\end{figure*} 

\begin{figure}[ht]
    \centering % <-- added
\begin{subfigure}{0.19\textwidth}
  \includegraphics[width=\linewidth]{./figures/results/trex.png}
\end{subfigure}
\begin{turn}{90} \; w/o CL \end{turn}
\begin{subfigure}{0.19\textwidth}
  \includegraphics[width=\linewidth]{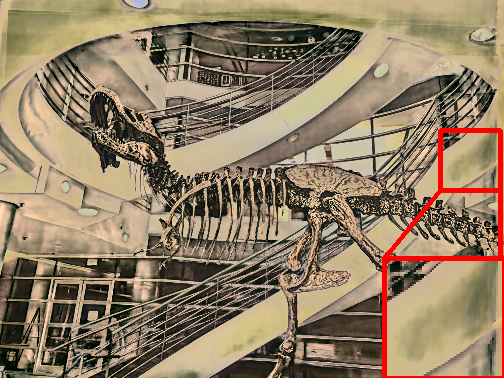}
\end{subfigure} % <-- added
\begin{subfigure}{0.19\textwidth}
  \includegraphics[width=\linewidth]{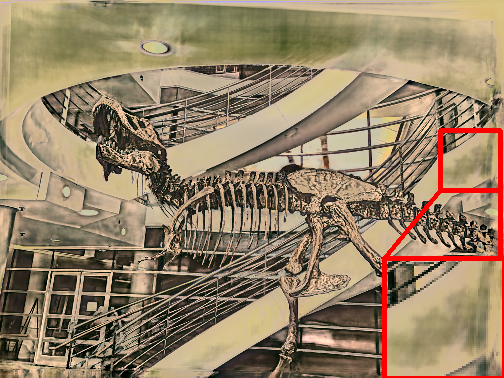}
\end{subfigure}
\begin{subfigure}{0.19\textwidth}
  \includegraphics[width=\linewidth]{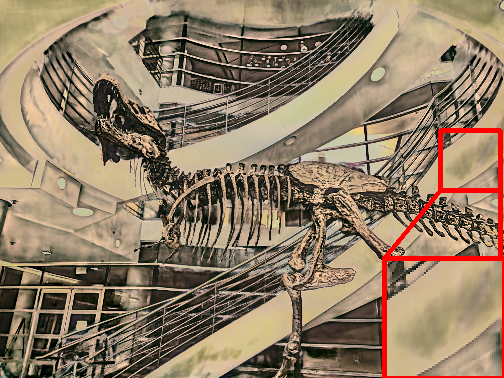}
\end{subfigure} % <-- added
\medskip

\begin{subfigure}{0.19\textwidth}
  \includegraphics[width=\linewidth]{./figures/style_images_same_res/oskar-kokoschka_not_detected_235932.jpg}
\end{subfigure}
\begin{turn}{90} \; \, w/ CL \end{turn}
\begin{subfigure}{0.19\textwidth}
  \includegraphics[width=\linewidth]{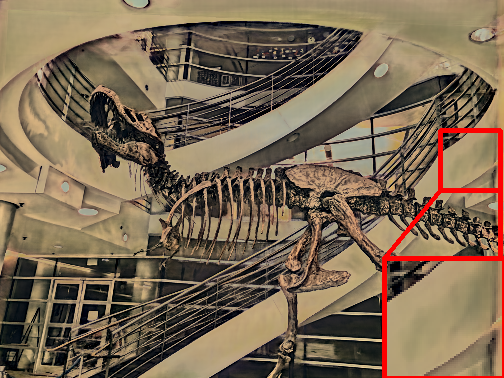}
\end{subfigure} % <-- added
\begin{subfigure}{0.19\textwidth}
  \includegraphics[width=\linewidth]{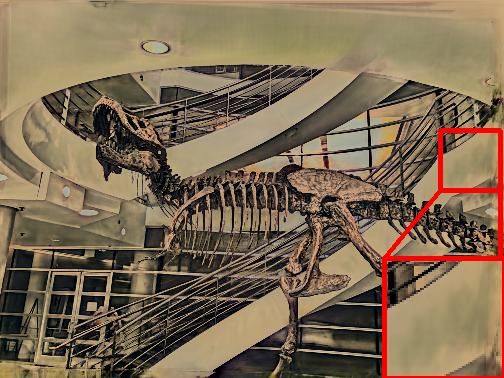}
\end{subfigure}
\begin{subfigure}{0.19\textwidth}
  \includegraphics[width=\linewidth]{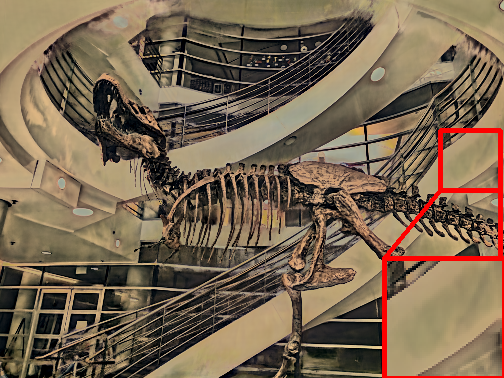}
\end{subfigure} % <-- added
\smallskip

\vspace{-0.3cm}
 --- --- --- --- ---  --- --- --- --- --- --- --- ---  --- --- --- --- --- --- --- --- --- 

\begin{subfigure}{0.19\textwidth}
  \includegraphics[width=\linewidth]{./figures/results/trex.png}
\end{subfigure}
\begin{turn}{90} \; w/o CL \end{turn}
\begin{subfigure}{0.19\textwidth}
  \includegraphics[width=\linewidth]{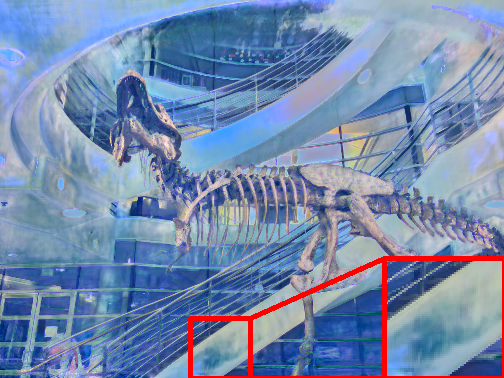}
\end{subfigure}
\begin{subfigure}{0.19\textwidth}
  \includegraphics[width=\linewidth]{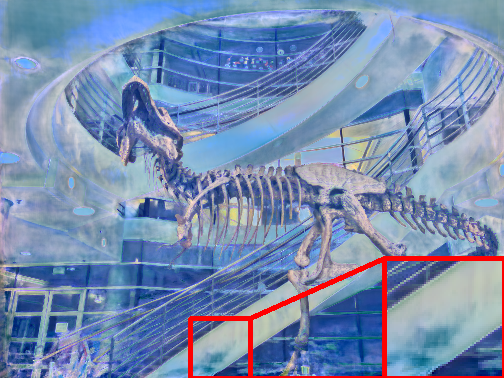}
\end{subfigure}
\begin{subfigure}{0.19\textwidth}
  \includegraphics[width=\linewidth]{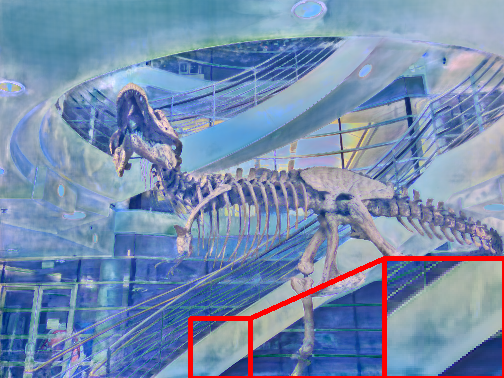}
\end{subfigure} % <-- added
\medskip

\begin{subfigure}{0.19\textwidth}
  \includegraphics[width=\linewidth]{./figures/style_images_same_res/pablo-picasso_the-family-of-blind-man-1903.jpg}
  \caption{Scene/Style}
\end{subfigure}
\begin{turn}{90} \quad \; \; w/ CL \end{turn}
\begin{subfigure}{0.19\textwidth}
  \includegraphics[width=\linewidth]{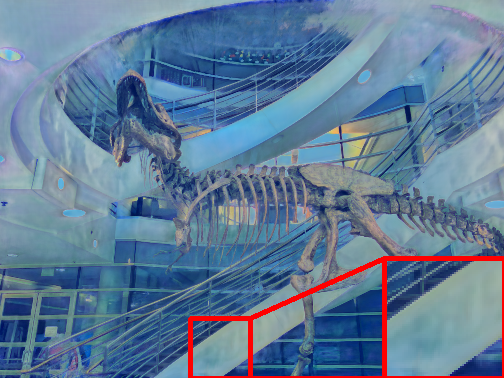}
  \caption{View 1}
\end{subfigure}
\begin{subfigure}{0.19\textwidth}
  \includegraphics[width=\linewidth]{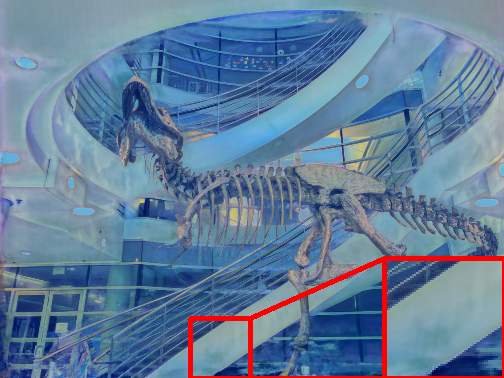}
  \caption{View 2}
\end{subfigure}
\begin{subfigure}{0.19\textwidth}
  \includegraphics[width=\linewidth]{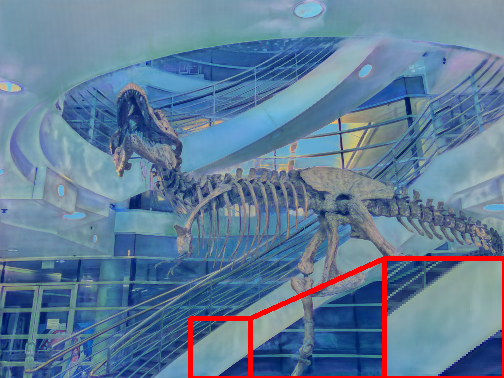}
  \caption{View 3}
\end{subfigure}
\smallskip

\caption{\textbf{Ablation study on Multi-View Consistency Loss.} Our proposed multi-view consistency loss effectively preserves content across different views, as demonstrated by the comparison between different views with and without the consistency loss. The consistent preservation of content shows the strength of our novel loss in 3D style transfer.}
\label{fig:consistency_ablation}
\end{figure}

\subsubsection{Time Comparison.}
 As shown in Tab. \ref{table:2}, StyleRF necessitates a total training of 5 hours and 38 minutes per-scene, along with an inference time of 18 seconds per frame. Hyper requires 4 days of training for the geometry learning phase and an additional 2 days of training for stylization, followed by a 50 seconds per-frame inference time. Our method only requires one training phase on top of the pretrained GNT network. After 1.5 days of single GPU training, our method successfully renders a stylized arbitrary novel view in only 21 seconds. 
 
 \subsubsection{Consistency Comparison.}
 we compare both long range and short range consistencies of our method with a consistency metric. We utilize the consistency metric defined in previous works \cite{Liu_2023_CVPR,chiang2022stylizing}. For short range consistency calculation, we render videos from style transferred novel views and calculate the optical flow for each frame using the optical flow estimation network PWCNet~\cite{Sun2018PWC-Net}, hence a different network as in our consistency loss to ensure a fair comparison. We warp the current frame to the next frame using the predicted flow, report the LPIPS \cite{zhang2018perceptual} and RMSE scores between the successive frames. For the calculation of long range consistency, we predict optical flow from the current frame (t) to the 7th successive frame (t+7) and report the corresponding scores. As shown in Tab. \ref{table:3}, our method outperforms existing methods in both long range and short range consistencies. Both Fig. \ref{fig:result} and Fig. \ref{fig:comparison_stylerf} approve the consistency metric visually and show our results are consistent.

\subsubsection{Qualitative Comparison.}
A qualitative comparison is shown in Fig. \ref{fig:comparison}. Hyper \cite{chiang2022stylizing} produces blurry images with artifacts. For example, in the flower scene, blue dots are clearly visible. Also, Hyper \cite{chiang2022stylizing} fails to capture stylistic details such as the neon tiger stylization result. StyleRF \cite{Liu_2023_CVPR} performs well at stylization and produces more consistent results compared to Hyper \cite{chiang2022stylizing}. However, it also struggles to preserve multi-view consistency: Fig. \ref{fig:comparison_stylerf} shows that StyleRF cannot preserve local details across views, leading to inconsistency. In case of stylization, our method is comparable to StyleRF \cite{Liu_2023_CVPR} and achieves better quality than Hyper \cite{chiang2022stylizing}. Fig. \ref{fig:result} and Fig. \ref{fig:comparison_stylerf} show that our results are  consistent across different views. Our method preserves content in a single rendering forward pass, without any per-scene optimization.

\subsubsection{User Study.}
3D style transfer is a considerably new topic, lacking well-established and predefined metrics. In addition, stylization quality is a subjective topic. Hence, we conducted a user study to assess the stylization quality and consistency of novel views from our work along with Hyper \cite{chiang2022stylizing} and StyleRF \cite{Liu_2023_CVPR}. In the survey, we presented scene and style image pairs along with randomly ordered results of our work and the related works. We asked the users to rate the stylization and consistency of the results on a scale of 1 to 5, with 1 being "Poor" and 5 being "Excellent". We collected a total of 150 votes per work. The results of the user study is stated in Tab. \ref{table:4}. As can be seen from the table, our method performs very similar to StyleRF \cite{Liu_2023_CVPR} and it clearly outperforms Hyper \cite{chiang2022stylizing} in the stylization and the consistency metric. 
StyleRF performs slightly better than our method, because they do per-scene overfitting, which gives them an advantage, while our method operates on novel scenes. 

\subsection{Ablation Study}

\subsubsection{Consistency Loss.} We conduct an ablation study to evaluate our proposed multi-view consistency loss in the context of 3D style transfer. To investigate the impact of the consistency loss on the preservation of multi-view consistency, \cref{fig:consistency_ablation} compares the results with and without the consistency loss. It shows that the consistency loss preserves colors corresponding to the same 3D points across different views. In contrast, without the consistency loss, the color of the same 3D points varies across viewpoints. Hence, when the consistency loss is applied, the generated outputs effectively preserve the multi-view consistency. This analysis demonstrates the importance of our proposed multi-view consistency loss in improving the quality and robustness of generalizable 3D style transfer.

\subsection{Limitations}
In our experiments, we show that we stylize an arbitrary scene with a given arbitrary style image at inference time, however, we think that there are possible future works and limitations of this work. 
We believe that it can be possible to further increase the stylization quality with the calculation of a style loss on local patches of the stylized image.
We further hypothesize that making the style loss depth-aware and angle-aware~\cite{hollein2022stylemesh}, could help to produce even more realistic stylizations of 3D scenes. 

\section{Conclusion}

In this paper, we present a novel generalizable 3D style transfer method that generalizes across both scenes and styles. Our method utilizes a generalizable NeRF method and a hypernetwork structure. We transform the intermediate features of a generalizable NeRF using a hypernetwork. We use an optical flow network to calculate the optical flow between source views and apply a novel consistency loss by warping the stylized images to the desired view. The pixels corresponding to the same point should have similar color values to minimize the consistency loss. Content, style, and multi-view consistency losses enable consistent stylization results across different viewpoints. After a single scene-independent pre-training, our framework is ready to use on novel scenes and styles immediately. In the end, our work enables a 3D style transfer on the fly by using source images and a single style image. We show the effectiveness of our framework by extensive qualitative and quantitative experiments. While other methods require hours of scene-specific training, our framework does not require any per-scene or per-style training. Our method produces high quality stylized images which preserve the geometric details and consistency across novel-views.
\bibliographystyle{splncs04}
\bibliography{main}

\begin{thebibliography}{10}
\providecommand{\url}[1]{\texttt{#1}}
\providecommand{\urlprefix}{URL }
\providecommand{\doi}[1]{https://doi.org/#1}

\bibitem{chan2022efficient}
Chan, E.R., Lin, C.Z., Chan, M.A., Nagano, K., Pan, B., De~Mello, S., Gallo, O., Guibas, L.J., Tremblay, J., Khamis, S., et~al.: Efficient geometry-aware 3d generative adversarial networks. In: Proceedings of the IEEE/CVF Conference on Computer Vision and Pattern Recognition. pp. 16123--16133 (2022)

\bibitem{tensorf}
Chen, A., Xu, Z., Geiger, A., Yu, J., Su, H.: Tensorf: Tensorial radiance fields. In: European Conference on Computer Vision (ECCV) (2022)

\bibitem{mvsnerf}
Chen, A., Xu, Z., Zhao, F., Zhang, X., Xiang, F., Yu, J., Su, H.: Mvsnerf: Fast generalizable radiance field reconstruction from multi-view stereo. In: Proceedings of the IEEE/CVF International Conference on Computer Vision. pp. 14124--14133 (2021)

\bibitem{chen2017coherent}
Chen, D., Liao, J., Yuan, L., Yu, N., Hua, G.: Coherent online video style transfer. In: Proceedings of the IEEE International Conference on Computer Vision. pp. 1105--1114 (2017)

\bibitem{chen2020optical}
Chen, X., Zhang, Y., Wang, Y., Shu, H., Xu, C., Xu, C.: Optical flow distillation: Towards efficient and stable video style transfer. In: European Conference on Computer Vision. pp. 614--630. Springer (2020)

\bibitem{chen2022upst}
Chen, Y., Yuan, Q., Li, Z., Xie, Y.L.W.W.C., Wen, X., Yu, Q.: Upst-nerf: Universal photorealistic style transfer of neural radiance fields for 3d scene. arXiv preprint arXiv:2208.07059  (2022)

\bibitem{chen2019learning}
Chen, Z., Zhang, H.: Learning implicit fields for generative shape modeling. In: Proceedings of the IEEE/CVF Conference on Computer Vision and Pattern Recognition. pp. 5939--5948 (2019)

\bibitem{chiang2022stylizing}
Chiang, P.Z., Tsai, M.S., Tseng, H.Y., Lai, W.S., Chiu, W.C.: Stylizing 3d scene via implicit representation and hypernetwork. In: Proceedings of the IEEE/CVF Winter Conference on Applications of Computer Vision. pp. 1475--1484 (2022)

\bibitem{SRF}
Chibane, J., Bansal, A., Lazova, V., Pons-Moll, G.: Stereo radiance fields (srf): Learning view synthesis from sparse views of novel scenes. In: {IEEE} Conference on Computer Vision and Pattern Recognition (CVPR). {IEEE} (jun 2021)

\bibitem{chiu2020iterative}
Chiu, T.Y., Gurari, D.: Iterative feature transformation for fast and versatile universal style transfer. In: European Conference on Computer Vision. pp. 169--184. Springer (2020)

\bibitem{mmnerf}
Dong, B., Chen, K., Wang, Z., Yan, M., Gu, J., Sun, X.: Mm-nerf: Large-scale scene representation with multi-resolution hash grid and multi-view priors features. Electronics  \textbf{13}(5) (2024)

\bibitem{downs2022google}
Downs, L., Francis, A., Koenig, N., Kinman, B., Hickman, R., Reymann, K., McHugh, T.B., Vanhoucke, V.: Google scanned objects: A high-quality dataset of 3d scanned household items. In: 2022 International Conference on Robotics and Automation (ICRA). pp. 2553--2560. IEEE (2022)

\bibitem{fan2022unified}
Fan, Z., Jiang, Y., Wang, P., Gong, X., Xu, D., Wang, Z.: Unified implicit neural stylization. In: European Conference on Computer Vision (2022)

\bibitem{flynn2019deepview}
Flynn, J., Broxton, M., Debevec, P., DuVall, M., Fyffe, G., Overbeck, R., Snavely, N., Tucker, R.: Deepview: View synthesis with learned gradient descent. In: Proceedings of the IEEE/CVF Conference on Computer Vision and Pattern Recognition. pp. 2367--2376 (2019)

\bibitem{fridovich2022plenoxels}
Fridovich-Keil, S., Yu, A., Tancik, M., Chen, Q., Recht, B., Kanazawa, A.: Plenoxels: Radiance fields without neural networks. In: Proceedings of the IEEE/CVF Conference on Computer Vision and Pattern Recognition. pp. 5501--5510 (2022)

\bibitem{gao2018reconet}
Gao, C., Gu, D., Zhang, F., Yu, Y.: Reconet: Real-time coherent video style transfer network. In: Asian Conference on Computer Vision. pp. 637--653. Springer (2018)

\bibitem{gao2020fast}
Gao, W., Li, Y., Yin, Y., Yang, M.H.: Fast video multi-style transfer. In: Proceedings of the IEEE/CVF Winter Conference on Applications of Computer Vision. pp. 3222--3230 (2020)

\bibitem{gatys2016image}
Gatys, L.A., Ecker, A.S., Bethge, M.: Image style transfer using convolutional neural networks. In: Proceedings of the IEEE conference on computer vision and pattern recognition. pp. 2414--2423 (2016)

\bibitem{gatys2017controlling}
Gatys, L.A., Ecker, A.S., Bethge, M., Hertzmann, A., Shechtman, E.: Controlling perceptual factors in neural style transfer. In: Proceedings of the IEEE Conference on Computer Vision and Pattern Recognition. pp. 3985--3993 (2017)

\bibitem{gupta2017characterizing}
Gupta, A., Johnson, J., Alahi, A., Fei-Fei, L.: Characterizing and improving stability in neural style transfer. In: Proceedings of the IEEE International Conference on Computer Vision (2017)

\bibitem{hollein2022stylemesh}
H{\"o}llein, L., Johnson, J., Nie{\ss}ner, M.: Stylemesh: Style transfer for indoor 3d scene reconstructions. In: Proceedings of the IEEE/CVF Conference on Computer Vision and Pattern Recognition. pp. 6198--6208 (2022)

\bibitem{huang2021learning}
Huang, H.P., Tseng, H.Y., Saini, S., Singh, M., Yang, M.H.: Learning to stylize novel views. In: Proceedings of the IEEE/CVF International Conference on Computer Vision. pp. 13869--13878 (2021)

\bibitem{huang2017arbitrary}
Huang, X., Belongie, S.: Arbitrary style transfer in real-time with adaptive instance normalization. In: Proceedings of the IEEE International Conference on Computer Vision. pp. 1501--1510 (2017)

\bibitem{Huang22StylizedNeRF}
Huang, Y.H., He, Y., Yuan, Y.J., Lai, Y.K., Gao, L.: Stylizednerf: Consistent 3d scene stylization as stylized nerf via 2d-3d mutual learning. In: Computer Vision and Pattern Recognition (CVPR) (2022)

\bibitem{johnson2016perceptual}
Johnson, J., Alahi, A., Fei-Fei, L.: Perceptual losses for real-time style transfer and super-resolution. In: European conference on computer vision. pp. 694--711. Springer (2016)

\bibitem{kingma2022autoencoding}
Kingma, D.P., Welling, M.: Auto-encoding variational bayes (2022)

\bibitem{li2018learning}
Li, X., Liu, S., Kautz, J., Yang, M.H.: Learning linear transformations for fast arbitrary style transfer. In: IEEE Conference on Computer Vision and Pattern Recognition (2019)

\bibitem{NIPS2017_49182f81}
Li, Y., Fang, C., Yang, J., Wang, Z., Lu, X., Yang, M.H.: Universal style transfer via feature transforms. In: Guyon, I., Luxburg, U.V., Bengio, S., Wallach, H., Fergus, R., Vishwanathan, S., Garnett, R. (eds.) Advances in Neural Information Processing Systems. vol.~30. Curran Associates, Inc. (2017)

\bibitem{Liu_2023_CVPR}
Liu, K., Zhan, F., Chen, Y., Zhang, J., Yu, Y., El~Saddik, A., Lu, S., Xing, E.P.: Stylerf: Zero-shot 3d style transfer of neural radiance fields. In: Proceedings of the IEEE/CVF Conference on Computer Vision and Pattern Recognition (CVPR). pp. 8338--8348 (June 2023)

\bibitem{liu2020neural}
Liu, L., Gu, J., Zaw~Lin, K., Chua, T.S., Theobalt, C.: Neural sparse voxel fields. Advances in Neural Information Processing Systems  \textbf{33},  15651--15663 (2020)

\bibitem{mescheder2019occupancy}
Mescheder, L., Oechsle, M., Niemeyer, M., Nowozin, S., Geiger, A.: Occupancy networks: Learning 3d reconstruction in function space. In: Proceedings of the IEEE/CVF conference on computer vision and pattern recognition. pp. 4460--4470 (2019)

\bibitem{miao2024conrf}
Miao, X., Bai, Y., Duan, H., Wan, F., Huang, Y., Long, Y., Zheng, Y.: Conrf: Zero-shot stylization of 3d scenes with conditioned radiation fields (2024)

\bibitem{Michel_2022_CVPR}
Michel, O., Bar-On, R., Liu, R., Benaim, S., Hanocka, R.: Text2mesh: Text-driven neural stylization for meshes. In: Proceedings of the IEEE/CVF Conference on Computer Vision and Pattern Recognition (CVPR). pp. 13492--13502 (June 2022)

\bibitem{mildenhall2019local}
Mildenhall, B., Srinivasan, P.P., Ortiz-Cayon, R., Kalantari, N.K., Ramamoorthi, R., Ng, R., Kar, A.: Local light field fusion: Practical view synthesis with prescriptive sampling guidelines. ACM Transactions on Graphics (TOG)  \textbf{38}(4),  1--14 (2019)

\bibitem{mildenhall2020nerf}
Mildenhall, B., Srinivasan, P.P., Tancik, M., Barron, J.T., Ramamoorthi, R., Ng, R.: Nerf: Representing scenes as neural radiance fields for view synthesis. In: European conference on computer vision. pp. 405--421. Springer (2020)

\bibitem{mu20223d}
Mu, F., Wang, J., Wu, Y., Li, Y.: 3d photo stylization: Learning to generate stylized novel views from a single image. In: Proceedings of the IEEE/CVF Conference on Computer Vision and Pattern Recognition. pp. 16273--16282 (2022)

\bibitem{mueller2022instant}
M\"uller, T., Evans, A., Schied, C., Keller, A.: Instant neural graphics primitives with a multiresolution hash encoding. ACM Trans. Graph.  \textbf{41}(4),  102:1--102:15 (Jul 2022)

\bibitem{nguyenphuoc2022snerf}
Nguyen-Phuoc, T., Liu, F., Xiao, L.: Snerf: stylized neural implicit representations for 3d scenes. ACM Trans. Graph.  \textbf{41}(4) (jul 2022)

\bibitem{niemeyer2020differentiable}
Niemeyer, M., Mescheder, L., Oechsle, M., Geiger, A.: Differentiable volumetric rendering: Learning implicit 3d representations without 3d supervision. In: Proceedings of the IEEE/CVF Conference on Computer Vision and Pattern Recognition. pp. 3504--3515 (2020)

\bibitem{ruder2016artistic}
Ruder, M., Dosovitskiy, A., Brox, T.: Artistic style transfer for videos. In: German conference on pattern recognition. pp. 26--36. Springer (2016)

\bibitem{ruder2018artistic}
Ruder, M., Dosovitskiy, A., Brox, T.: Artistic style transfer for videos and spherical images. International Journal of Computer Vision  \textbf{126}(11),  1199--1219 (2018)

\bibitem{saleh2015large}
Saleh, B., Elgammal, A.: Large-scale classification of fine-art paintings: Learning the right metric on the right feature. arXiv preprint arXiv:1505.00855  (2015)

\bibitem{kplanes_2023}
{Sara Fridovich-Keil and Giacomo Meanti}, Warburg, F.R., Recht, B., Kanazawa, A.: K-planes: Explicit radiance fields in space, time, and appearance. In: CVPR (2023)

\bibitem{simonyan2014very}
Simonyan, K., Zisserman, A.: Very deep convolutional networks for large-scale image recognition. arXiv preprint arXiv:1409.1556  (2014)

\bibitem{sun2022direct}
Sun, C., Sun, M., Chen, H.T.: Direct voxel grid optimization: Super-fast convergence for radiance fields reconstruction. In: Proceedings of the IEEE/CVF Conference on Computer Vision and Pattern Recognition. pp. 5459--5469 (2022)

\bibitem{Sun2018PWC-Net}
Sun, D., Yang, X., Liu, M.Y., Kautz, J.: {PWC-Net}: {CNNs} for optical flow using pyramid, warping, and cost volume. In: CVPR (2018)

\bibitem{Svoboda_2020_CVPR}
Svoboda, J., Anoosheh, A., Osendorfer, C., Masci, J.: Two-stage peer-regularized feature recombination for arbitrary image style transfer. In: Proceedings of the IEEE/CVF Conference on Computer Vision and Pattern Recognition (CVPR) (June 2020)

\bibitem{t2023is}
T, M.V., Wang, P., Chen, X., Chen, T., Venugopalan, S., Wang, Z.: Is attention all that ne{RF} needs? In: The Eleventh International Conference on Learning Representations (2023)

\bibitem{teed2020raft}
Teed, Z., Deng, J.: Raft: Recurrent all-pairs field transforms for optical flow. In: European conference on computer vision. pp. 402--419. Springer (2020)

\bibitem{ulyanov2016texture}
Ulyanov, D., Lebedev, V., Vedaldi, A., Lempitsky, V.S.: Texture networks: Feed-forward synthesis of textures and stylized images. In: ICML (2016)

\bibitem{wang2021ibrnet}
Wang, Q., Wang, Z., Genova, K., Srinivasan, P.P., Zhou, H., Barron, J.T., Martin-Brualla, R., Snavely, N., Funkhouser, T.: Ibrnet: Learning multi-view image-based rendering. In: Proceedings of the IEEE/CVF Conference on Computer Vision and Pattern Recognition. pp. 4690--4699 (2021)

\bibitem{ReReVST2020}
Wang, W., Yang, S., Xu, J., Liu, J.: Consistent video style transfer via relaxation and regularization. {IEEE} Trans. Image Process.  (2020)

\bibitem{xia2021real}
Xia, X., Xue, T., Lai, W.s., Sun, Z., Chang, A., Kulis, B., Chen, J.: Real-time localized photorealistic video style transfer. In: Proceedings of the IEEE/CVF Winter Conference on Applications of Computer Vision. pp. 1089--1098 (2021)

\bibitem{yariv2020multiview}
Yariv, L., Kasten, Y., Moran, D., Galun, M., Atzmon, M., Ronen, B., Lipman, Y.: Multiview neural surface reconstruction by disentangling geometry and appearance. Advances in Neural Information Processing Systems  \textbf{33},  2492--2502 (2020)

\bibitem{yin20213dstylenet}
Yin, K., Gao, J., Shugrina, M., Khamis, S., Fidler, S.: 3dstylenet: Creating 3d shapes with geometric and texture style variations. In: Proceedings of the IEEE/CVF International Conference on Computer Vision. pp. 12456--12465 (2021)

\bibitem{yu2021pixelnerf}
Yu, A., Ye, V., Tancik, M., Kanazawa, A.: pixelnerf: Neural radiance fields from one or few images (2021)

\bibitem{zhang2022arf}
Zhang, K., Kolkin, N., Bi, S., Luan, F., Xu, Z., Shechtman, E., Snavely, N.: Arf: Artistic radiance fields. In: Computer Vision--ECCV 2022: 17th European Conference, Tel Aviv, Israel, October 23--27, 2022, Proceedings, Part XXXI. pp. 717--733. Springer (2022)

\bibitem{zhang2018perceptual}
Zhang, R., Isola, P., Efros, A.A., Shechtman, E., Wang, O.: The unreasonable effectiveness of deep features as a perceptual metric. In: CVPR (2018)

\bibitem{zhou2018stereo}
Zhou, T., Tucker, R., Flynn, J., Fyffe, G., Snavely, N.: Stereo magnification: Learning view synthesis using multiplane images. arXiv preprint arXiv:1805.09817  (2018)

\end{thebibliography}

\section{Architecture Details}
\begin{figure}[htb]
  \centering
  \vspace{-0.8cm}
  \includegraphics[width=0.58\textwidth]{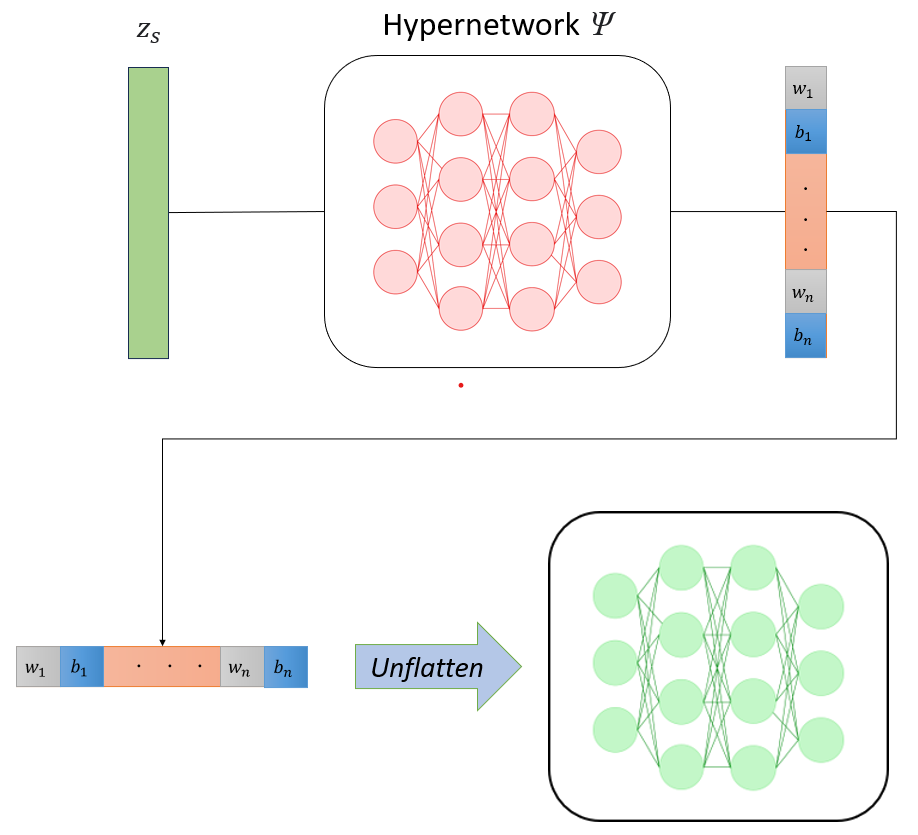}
  \vspace{-0.2cm}
  \caption{\textbf{Hypernetwork.} Our hypernetwork generates weights and biases for an intermediate MLP which is used during rendering to apply the feature transformation leading to a stylized output view.
  }
  \vspace{-1.0cm}
  \label{fig:hypernet}
\end{figure}
\begin{figure}
    \centering % <-- added
    \begin{subfigure}{0.18\textwidth}
  \includegraphics[width=\linewidth]{./figures/results/fern/image001.png}
\end{subfigure}
\begin{subfigure}{0.18\textwidth}
  \includegraphics[width=\linewidth]{./figures/style_images_same_res/landscape-under-a-stormy-sky-1888.jpg}
\end{subfigure} % <-- added
\begin{subfigure}{0.18\textwidth}
  \includegraphics[width=\linewidth]{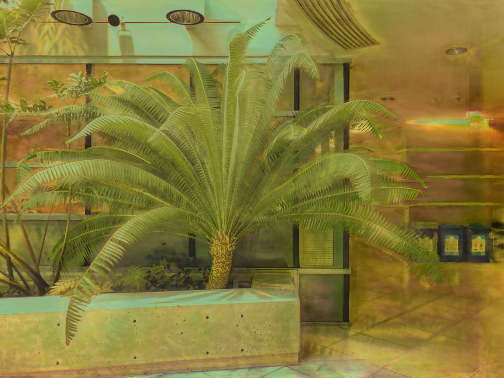}
\end{subfigure}
\begin{subfigure}{0.18\textwidth}
  \includegraphics[width=\linewidth]{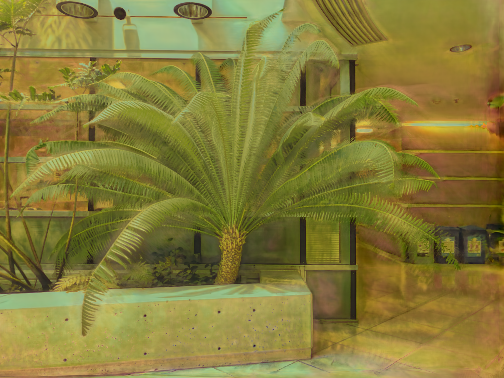}
\end{subfigure} % <-- added
\begin{subfigure}{0.18\textwidth}
  \includegraphics[width=\linewidth]{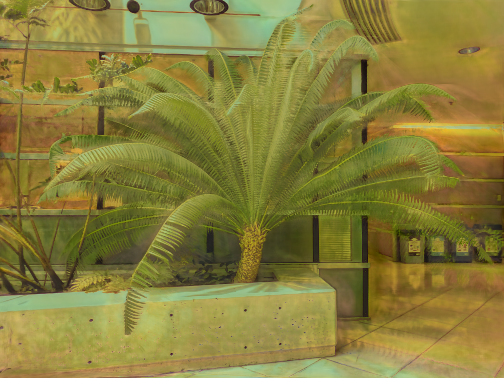}
\end{subfigure} % <-- added
\medskip

\begin{subfigure}{0.18\textwidth}
  \includegraphics[width=\linewidth]{./figures/results/fern/image001.png}
\end{subfigure}
\begin{subfigure}{0.18\textwidth}
  \includegraphics[width=\linewidth]{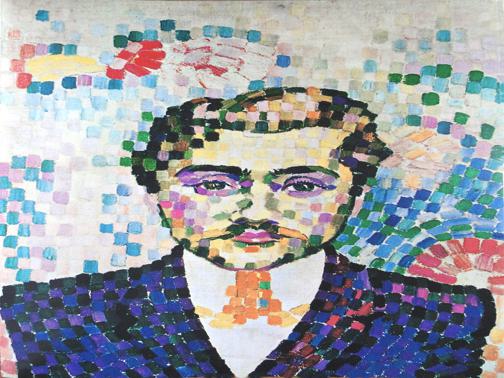}
\end{subfigure} % <-- added
\begin{subfigure}{0.18\textwidth}
  \includegraphics[width=\linewidth]{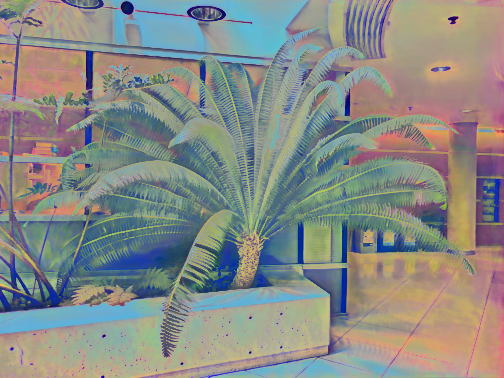}
\end{subfigure}
\begin{subfigure}{0.18\textwidth}
  \includegraphics[width=\linewidth]{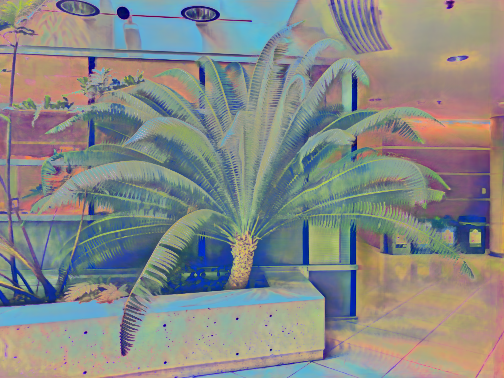}
\end{subfigure} % <-- added
\begin{subfigure}{0.18\textwidth}
  \includegraphics[width=\linewidth]{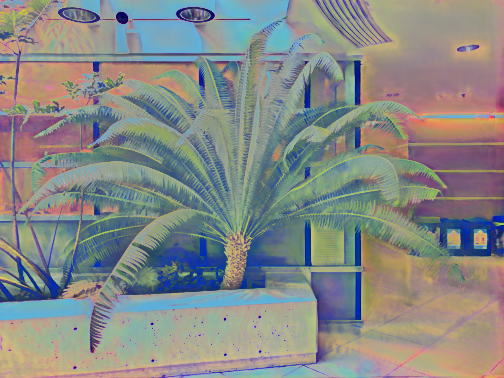}
\end{subfigure} % <-- added
\medskip

\begin{subfigure}{0.18\textwidth}
  \includegraphics[width=\linewidth]{./figures/results/fern/image001.png}
\end{subfigure}
\begin{subfigure}{0.18\textwidth}
  \includegraphics[width=\linewidth]{./figures/style_images_same_res/the-ill-matched-couple.jpg}
\end{subfigure} % <-- added
\begin{subfigure}{0.18\textwidth}
  \includegraphics[width=\linewidth]{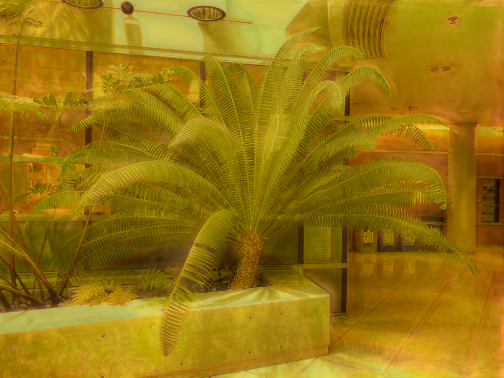}
\end{subfigure}
\begin{subfigure}{0.18\textwidth}
  \includegraphics[width=\linewidth]{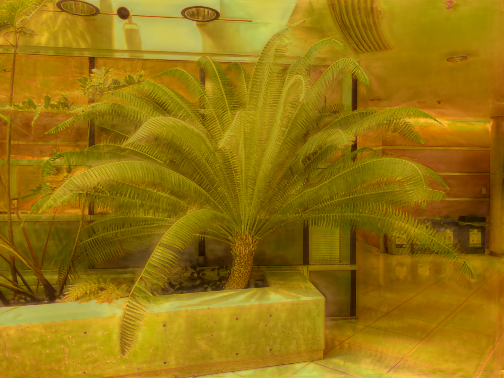}
\end{subfigure} % <-- added
\begin{subfigure}{0.18\textwidth}
  \includegraphics[width=\linewidth]{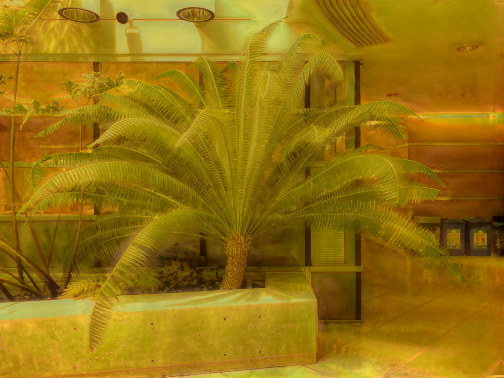}
\end{subfigure} % <-- added
\medskip

\begin{subfigure}{0.18\textwidth}
  \includegraphics[width=\linewidth]{./figures/results/orchids/image000.png}
\end{subfigure}
\begin{subfigure}{0.18\textwidth}
  \includegraphics[width=\linewidth]{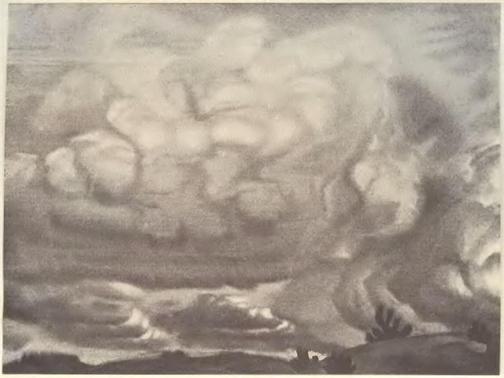}
\end{subfigure} % <-- added
\begin{subfigure}{0.18\textwidth}
  \includegraphics[width=\linewidth]{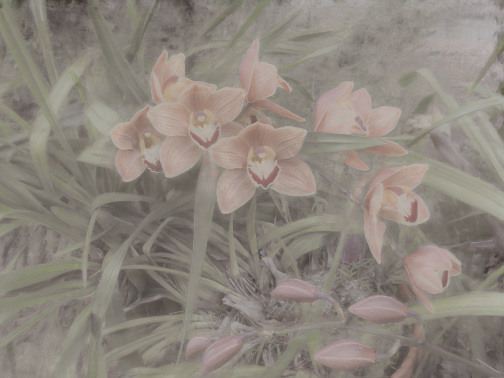}
\end{subfigure}
\begin{subfigure}{0.18\textwidth}
  \includegraphics[width=\linewidth]{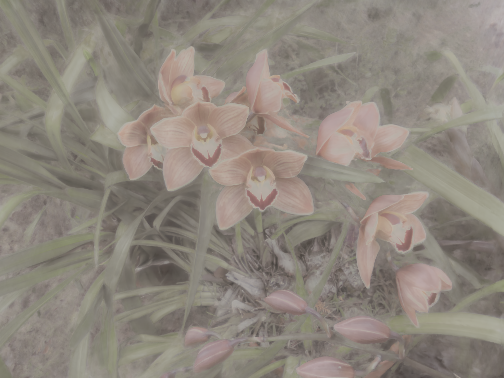}
\end{subfigure} % <-- added
\begin{subfigure}{0.18\textwidth}
  \includegraphics[width=\linewidth]{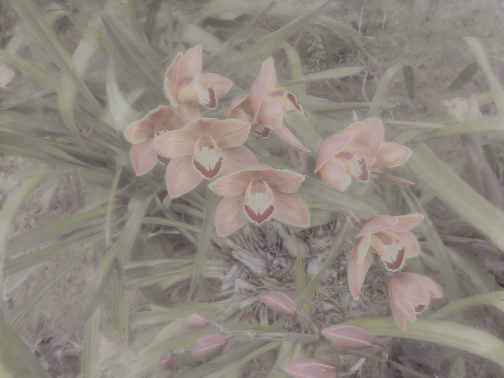}
\end{subfigure} % <-- added
\medskip

\begin{subfigure}{0.18\textwidth}
  \includegraphics[width=\linewidth]{./figures/results/orchids/image000.png}
\end{subfigure}
\begin{subfigure}{0.18\textwidth}
  \includegraphics[width=\linewidth]{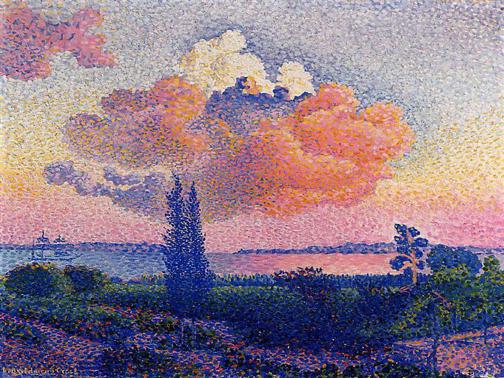}
\end{subfigure} % <-- added
\begin{subfigure}{0.18\textwidth}
  \includegraphics[width=\linewidth]{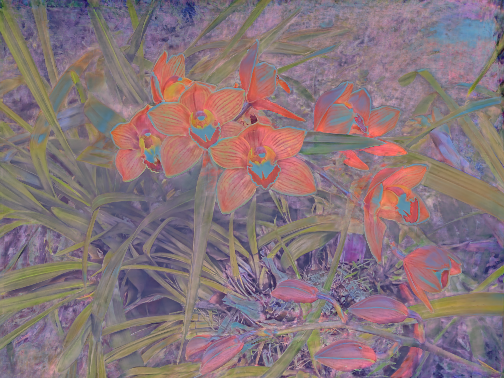}
\end{subfigure}
\begin{subfigure}{0.18\textwidth}
  \includegraphics[width=\linewidth]{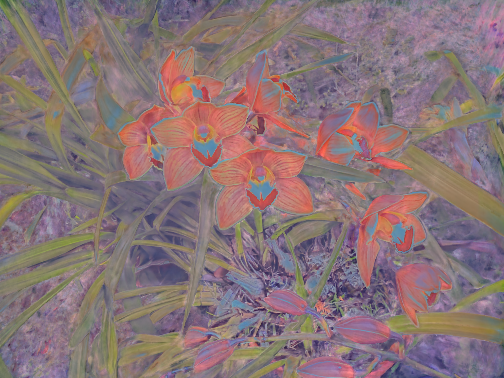}
\end{subfigure} % <-- added
\begin{subfigure}{0.18\textwidth}
  \includegraphics[width=\linewidth]{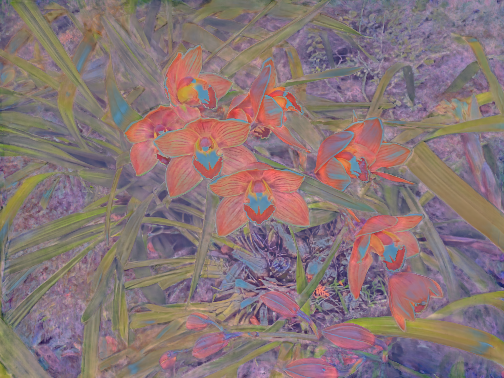}
\end{subfigure} % <-- added
\medskip

\begin{subfigure}{0.18\textwidth}
  \includegraphics[width=\linewidth]{./figures/results/orchids/image000.png}
\end{subfigure}
\begin{subfigure}{0.18\textwidth}
  \includegraphics[width=\linewidth]{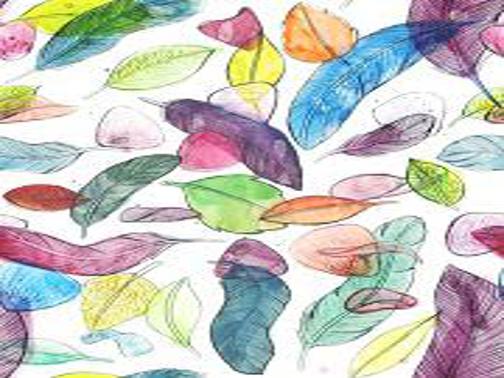}
\end{subfigure} % <-- added
\begin{subfigure}{0.18\textwidth}
  \includegraphics[width=\linewidth]{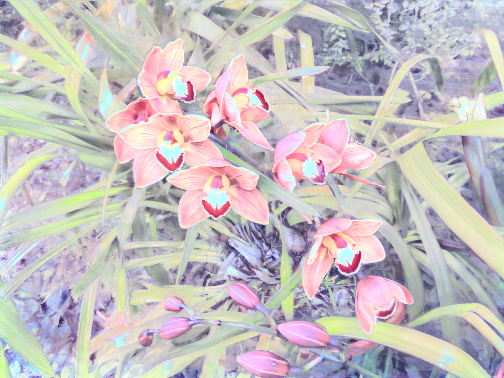}
\end{subfigure}
\begin{subfigure}{0.18\textwidth}
  \includegraphics[width=\linewidth]{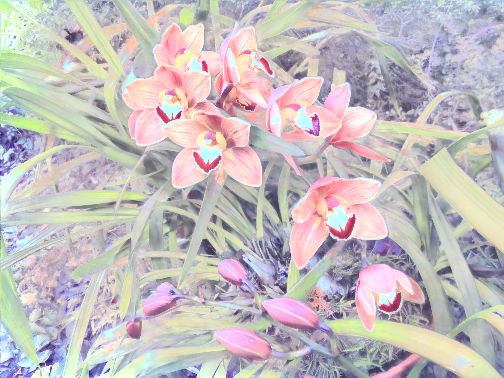}
\end{subfigure} % <-- added
\begin{subfigure}{0.18\textwidth}
  \includegraphics[width=\linewidth]{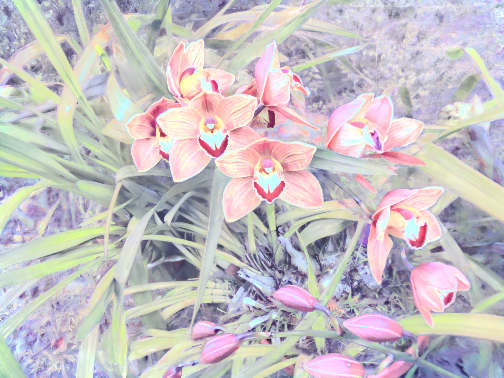}
\end{subfigure} % <-- added
\medskip

\begin{subfigure}{0.18\textwidth}
  \includegraphics[width=\linewidth]{./figures/results/room.png}
\end{subfigure}
\begin{subfigure}{0.18\textwidth}
  \includegraphics[width=\linewidth]{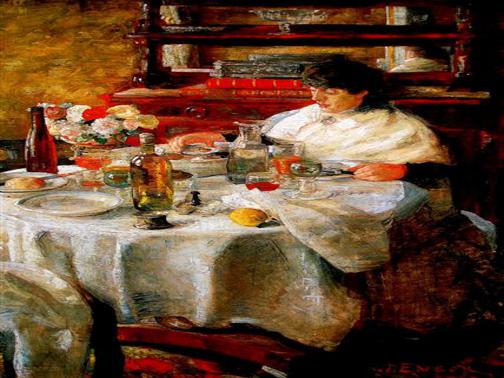}
\end{subfigure}
\begin{subfigure}{0.18\textwidth}
  \includegraphics[width=\linewidth]{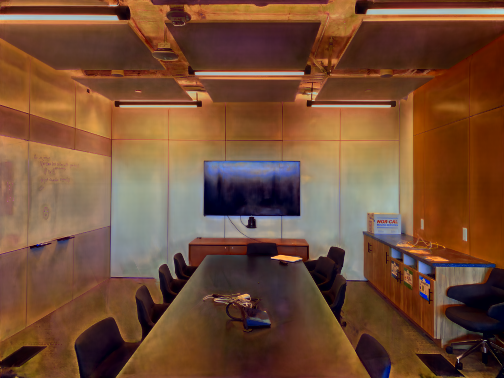}
\end{subfigure}
\begin{subfigure}{0.18\textwidth}
  \includegraphics[width=\linewidth]{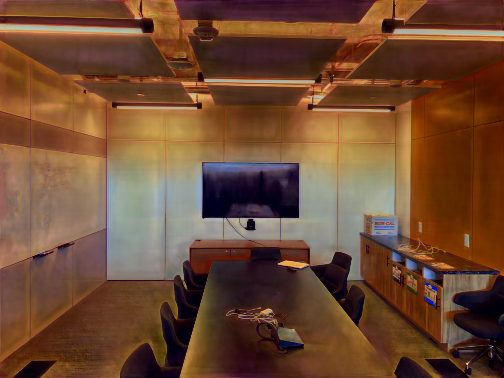}
\end{subfigure}
\begin{subfigure}{0.18\textwidth}
  \includegraphics[width=\linewidth]{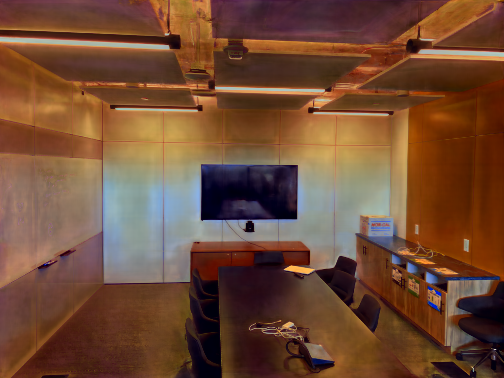}
\end{subfigure}
\medskip

\begin{subfigure}{0.18\textwidth}
  \includegraphics[width=\linewidth]{./figures/results/room.png}
\end{subfigure}
\begin{subfigure}{0.18\textwidth}
  \includegraphics[width=\linewidth]{./figures/style_images_same_res/pablo-picasso_the-family-of-blind-man-1903.jpg}
\end{subfigure}
\begin{subfigure}{0.18\textwidth}
  \includegraphics[width=\linewidth]{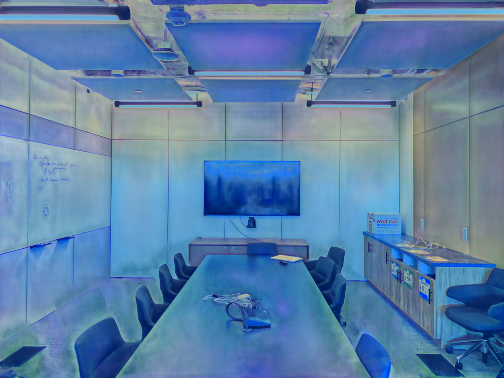}
\end{subfigure}
\begin{subfigure}{0.18\textwidth}
  \includegraphics[width=\linewidth]{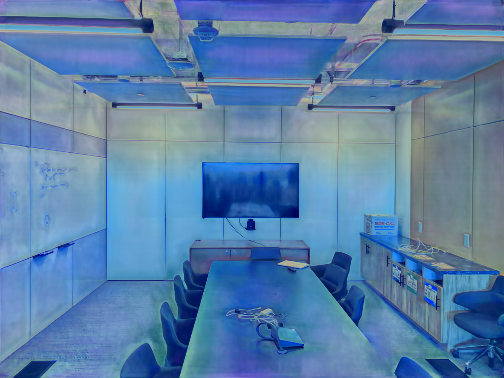}
\end{subfigure}
\begin{subfigure}{0.18\textwidth}
  \includegraphics[width=\linewidth]{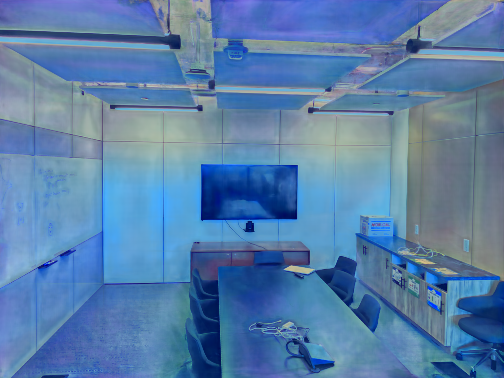}
\end{subfigure}
\medskip

\begin{subfigure}{0.18\textwidth}
  \includegraphics[width=\linewidth]{./figures/results/flower.png}
  \caption{Scene}
\end{subfigure}
\begin{subfigure}{0.18\textwidth}
  \includegraphics[width=\linewidth]{./figures/style_images_same_res/pablo-picasso_manola.jpg}
  \caption{Style}
\end{subfigure}
\begin{subfigure}{0.18\textwidth}
  \includegraphics[width=\linewidth]{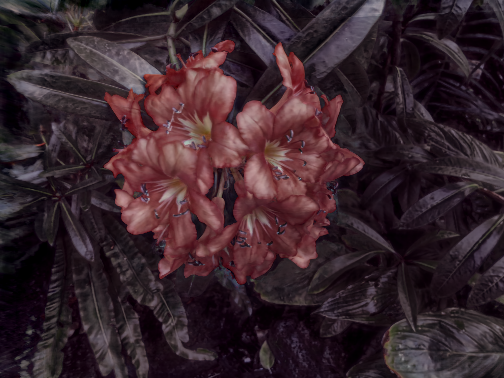}
  \caption{View 1}
\end{subfigure}
\begin{subfigure}{0.18\textwidth}
  \includegraphics[width=\linewidth]{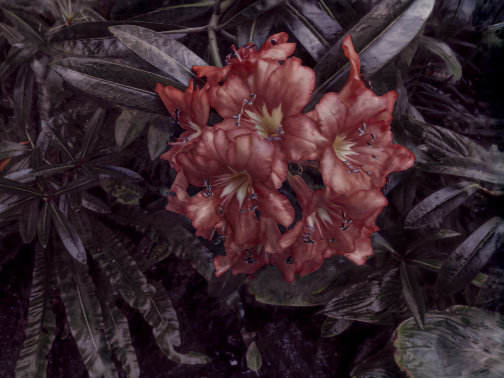}
  \caption{View 2}
\end{subfigure}
\begin{subfigure}{0.18\textwidth}
  \includegraphics[width=\linewidth]{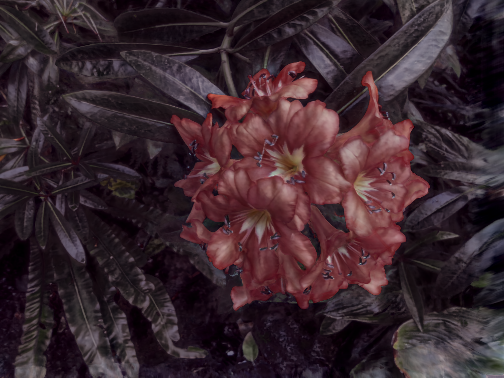}
  \caption{View 3}
\end{subfigure}
\smallskip

\caption{\textbf{Additional results.} We provide additional stylization results on different novel views and show that efficient stylization works on different scenes and style images.}
\label{fig:additional_results}
\end{figure}
\subsection{Style-VAE} For the extraction of style latent codes from the given style image, we follow the same architecture as \cite{chiang2022stylizing}. Style-VAE is a variational auto encoder (VAE \cite{kingma2022autoencoding}) based architecture to generate style images, it consists of an encoder ($E_{VAE}$) and a decoder ($D_{VAE}$). The VAE is trained on style images following the same VAE objective \cite{kingma2022autoencoding}. $E_{VAE}$ outputs a 1024 dimensional vector, where the first half of the elements contain the mean of the style latent ($\mu_{style}$) and the second half is the standard deviation ($\sigma_{style}$). We only use the $\mu_{style}$ as a style latent vector.

\subsection{NeRF with Hypernetwork} As shown in Fig. \ref{fig:hypernet}, our hypernetwork ($\psi$) takes a 512 dimensional style latent vector and outputs weights and biases of the intermediate MLP. Our hypernetwork architecture is a two layered MLP consisting of 1 hidden layer with 64 neurons. The hypernetwork outputs all weights and biases for the intermediate MLP. The intermediate MLP is also a two layered MLP with 1 hidden layer with 128 neurons. Given a 64 dimensional ray feature, it outputs a 64 dimensional stylized ray feature. For the view transformer and ray transformer, we follow the architecture of \cite{t2023is}.  

\section{Additional Qualitative Results}
In Fig. \ref{fig:additional_results}, we provide additional stylization results with various scenes and styles that were not seen during training. %from our validation dataset. 
We show that given an input style image, our method can stylize arbitrary scenes at inference time without requiring per-scene or per-style optimization. To indicate the preservation of multi-view consistency after the stylization we show the rendered results from different novel views. 

\clearpage
\end{document}